\documentclass[acmtog]{acmart}

\usepackage{booktabs} 
\usepackage{xcolor}

\usepackage[ruled]{algorithm2e} 

\SetAlFnt{\small}
\SetAlCapFnt{\small}
\SetAlCapNameFnt{\small}
\SetAlCapHSkip{0pt}
\DeclareMathOperator*{\argmin}{arg\,min}

\setcopyright{rightsretained}
\acmJournal{TOG}

\acmYear{2021}\acmVolume{40}\acmNumber{4}\acmArticle{148}\acmMonth{8}
\acmDOI{10.1145/3450626.3459871}

\begin{document}

\title{Consistent Depth of Moving Objects in Video}

\author{Zhoutong Zhang}
\affiliation{
  \institution{MIT CSAIL, Google Research}}

 \email{ztzhang@mit.edu}
\author{Forrester Cole}
\affiliation{%
  \institution{Google Research}}

\email{fcole@google.com}

\author{Richard Tucker}
\affiliation{%
  \institution{Google Research}}

\email{richardt@google.com}
\author{William T. Freeman}
\affiliation{%
  \institution{MIT CSAIL, Google Research}}

\email{billf@mit.edu}
\author{Tali Dekel}
\affiliation{
 \institution{Google Research, Weizmann Institute of Science}}

\email{tali.dekel@weizmann.ac.il}

\begin{abstract}

We present a method to estimate depth of a dynamic scene, containing arbitrary moving objects, from an ordinary video captured with a moving camera.  We seek a \emph{geometrically and temporally consistent} solution to this underconstrained problem: the depth predictions of corresponding points across frames should induce plausible, smooth motion in 3D. We formulate this objective in a new test-time training framework where a depth-prediction CNN is trained in tandem with an auxiliary scene-flow prediction MLP over the entire input video. By recursively unrolling the scene-flow prediction MLP over varying time steps, we compute both short-range scene flow to impose local smooth motion priors directly in 3D, and long-range scene flow to impose multi-view consistency constraints with wide baselines. We demonstrate accurate and temporally coherent results on a variety of challenging videos containing diverse moving objects (pets, people, cars), as well as camera motion. Our depth maps give rise to a number of depth-and-motion aware video editing effects such as object and lighting insertion.

\end{abstract}

%
%


%
%

\vspace{-10mm}
\begin{teaserfigure}
  \centering
    \includegraphics[width=0.96\linewidth]{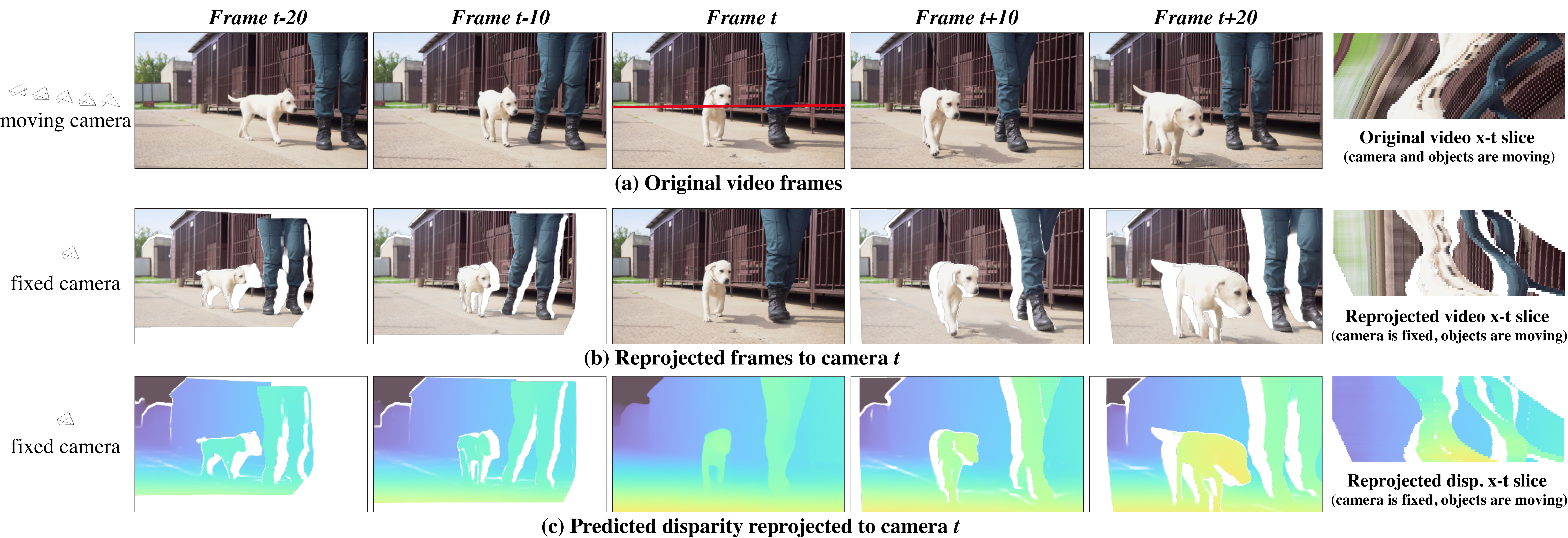}
    \vspace{-2mm}
    \captionof{figure}{Our method estimates \emph{geometrically and temporally consistent} depth from a general video containing fast-moving objects and camera motion. The input video (a) contains  a continuous camera dolly motion following the moving person and puppy. This is a difficult case for depth estimation due to the correlated motion between camera and subject. The video is shown reprojected into the camera at frame $t$ using our predicted depth (b), with disparity maps of the re-projection shown below (c). On the right: $x-t$ slices for the horizontal line marked in red on frame $t$ in (a). The slice of the original video (top) shows both camera and objects' motion (slanted lines in the background, twisted lines in the foreground). The slice of the re-projected frames (middle) shows the camera fixed relative to the background (vertical lines), and foreground objects moving relative to the camera (twisted lines).
    }
  \label{fig:teaser}
\end{teaserfigure}


\maketitle

\section{Introduction}

Estimating the geometry of moving objects from \emph{ordinary videos}---where both the camera and objects in the scene are moving---is an open, underconstrained problem. For any video of a moving object, there is a space of solutions that satisfy the visual evidence: the object could be far away and moving fast, or close and moving slowly. Thus, prior knowledge about objects' motion or shape is necessary to estimate geometry of moving objects in video.

Learning geometric priors directly from data has proven to be a promising approach that  facilitates geometry estimation in unconstrained setups such as predicting depth  from  a single RGB image (RGB-to-depth), or from a video where both the camera and the people in the scene are moving. In such cases, epipolar geometry constraints break and triangulation-based methods are not applicable, so the solution is determined by learned geometric priors. Such geometric priors are typically learned from a large dataset containing visual data (images/videos), and their corresponding depth maps, which are used as supervision during training.
While powerful, a \emph{purely data-driven} approach has two main  weaknesses in the case of arbitrary moving objects in a monocular video: (i) \emph{scarcity of training data}---video datasets of dynamic scenes with ground truth depth are still limited and only available for specific class of objects (e.g., moving people~\cite{li2020mannequinchallenge, li2019learning}), (ii) \emph{temporal consistency}---feed-forward models for videos consider a fixed, short-range window of inputs (often one or two frames), resulting in inconsistent geometry and flickering depth over time. 

Instead of relying entirely on data, we take a hybrid approach and initialize a solution from geometric priors learned from a large dataset, then optimize the solution to satisfy space-time geometric constraints.  This approach is inspired by the recent work of \cite{Luo-VideoDepth-2020}, which finetunes a monocular depth prediction model for a video by assuming a static scene and requiring depth estimates for corresponding pixels to be consistent across frames. Our method generalizes this approach to videos of \emph{dynamic scenes} by explicitly modeling \emph{scene flow}, i.e., the 3D motion of objects in the scene.

We seek a geometrically and temporally consistent solution: the depth and scene flow at corresponding points should to be consistent over frames, and the scene flow should change smoothly in time. Our key contribution is a formulation of this ill-posed problem via the optimization of two networks: a \emph{depth-prediction} CNN that takes an RGB frame as input and outputs a dense depth map; and an auxiliary \emph{scene-flow-prediction} multi-layer perceptron (MLP) that takes a 3D point as input and outputs its 3D velocity vector. The depth network is first initialized using a data-driven prior (pretrained weights), and then finetuned in tandem with the scene-flow network for a given input video, using a smooth-motion prior and multi-view consistency losses. As in \cite{Luo-VideoDepth-2020}, we compute camera poses and optical flow between pairs of frames in a pre-processing step and use them as input.

Our scene-flow MLP plays two important roles: (1) it provides an explicit, 3D representation of scene flow that aggregates information over space and time through the shared weights of the network, and produces plausible flow in cases where an analytic solution derived from depth and optical flow is unstable (e.g., nearly parallel rays between two points, as shown in Section~\ref{sec:cube_analysis}.).
(2) it provides a mechanism to form scene flow estimates over varying time steps. As demonstrated by stereo and MVS methods, wide baseline correspondences are required to accurately estimate depth, but smooth-motion priors must be evaluated locally. The scene-flow MLP, which maps a 3D point at time $t$ to its velocity vector (3D offset between $t$ and $t+1$), can be evaluated recursively by unrolling. This allows us to compute long-range scene flow to apply multi-view consistency losses between distant frames, and short-range scene flow to apply a smooth motion prior. 
 
We demonstrate detailed, accurate, and temporally-consistent depth maps produced from ordinary videos with multiple, non-rigid, rapidly-moving objects (Fig.~\ref{fig:teaser}). We evaluate and ablate our method quantitatively and show 40--45\% reduction in $L_1$ relative error vs. single-view depth prediction, and show various depth and motion aware editing effects in videos made possible by our depth and scene flow estimates.

\section{Related Work}
\paragraph{Learning-based depth prediction.} 

Deep learning based methods for predicting depth under different setups have demonstrated a dramatic improvement in recent years. 
Numerous methods have been proposed to predict depth from a single RGB image. Such methods typically use a feed-forward CNN-based model and can be divided into two main categories: \emph{supervised methods} that regress to  ground-truth depth maps given a large dataset of images and their corresponding depth maps~\cite{Ranftl2020,li2018megadepth,chen2016single,xian2018monocular,fu2018deep, eigen2014depth,wang2019web}, and \emph{self-supervised methods}~\cite{casser2019unsupervised,monodepth2,zhou2017unsupervised,godard2017unsupervised,yin2018geonet, yang2018every} that learn depth prediction by optimizing for photometric consistency, either using stereo images or monocular video. To handle moving objects in training videos, self-supervised methods predict an optical flow field as an additional task (e.g. \cite{yin2018geonet,yang2018every}), or learn to mask out moving objects (e.g. \cite{monodepth2}).

Once trained, single-frame methods are agnostic to moving objects, and can be applied to any video in a frame-by-frame fashion. However, because depth is predicted independently per-frame, such an approach results in temporally inconsistent estimates. The loss of epipolar constraints also means single-view prediction lags in accuracy over multi-view stereo (MVS) methods~\cite{seitz2006comparison,schoenberger2016mvs} for stationary regions.

\paragraph{Geometry estimation of dynamic scenes}

Methods for estimating geometry of a dynamic scene roughly fall into two major categories. The first category of works aim to solve this problem by considering either a multi-camera setup where epipolar geometry constrains can be applied, or using RGBD data (e.g., \cite{newcombe2015dynamicfusion, Dou2016Fusion4DRP, Innmann2016VolumeDeformRV, wedel2011stereoscopic, basha2012structure, basha2013multi, richardt2016dense, bansal20204d}). The second category of works aim to tackle this problem using monocular videos, which is a more challenging and ill-posed task. To deal with such challenges, some approaches aim to reconstruct the geometry of a dynamic scene by limiting in the type of scenes and objects' motion~\cite{ranftl2016dense, russell2014video, Park20103DRO, Simon2017KroneckerMarkovPF}. 

Other methods combine data-driven priors with multi-view stereo methods~\cite{li2019learning, rematas2018soccer,yoon2020novel}.  For example, \cite{li2019learning} uses parallax between two RGB frames to estimate depth in static parts of the image, then inpaints the depth of dynamic regions using a depth-prediction network.

In contrast, our method optimizes a pre-trained depth prediction network over the entire video at test-time to produce consistent estimations. 
The fusion method of \cite{yoon2020novel} uses the entire video to compute MVS estimates that are combined with a single-view CNN prediction using a learned module. Our method does not require MVS depth as input, only camera poses, and does not require training a fusion module. 
In addition, some methods ~\cite{klingner2020self,patil2020don} use semantic information to guide depth prediction and identify moving objects during training. Those methods usually focuses on specific scenarios such as autonomous driving. Our method does not require semantic information and aims to solve for consistent depth maps for general videos.

\paragraph{Test-time training for depth estimation}
Recently, \emph{test-time training} methods for depth estimation~\cite{chen2019self, casser2019depth, Luo-VideoDepth-2020} have appeared, which use deep neural networks as optimizers: given an objective function defined over the test data, the variables of optimization are the weights of a deep network that predicts the unknowns, rather than the unknowns themselves. By finetuning a pre-trained network on a single video, these methods both apply data-driven depth priors and aggregate information across the entire video.
The method of \cite{casser2019depth} segments the video and proposes separate motion models for each segment, whereas our scene-flow MLP learns a global motion model. Compared to GLNet~\cite{chen2019self}, we explicitly model 3D motion, rather than deriving it from optical flow and depth. We show that this explicit model of motion is important for stability and long-range correspondence. Most closely related to our work is \cite{Luo-VideoDepth-2020}, which shares our goal of accurate, temporally-consistent depth from video. While their method tolerates small amounts of object motion, it fundamentally assumes a stationary scene, whereas our method is designed to handle large object motion.

\paragraph{Reconstruction from smooth motion priors and non-rigid SfM} Prior to the rise of data-driven methods, reconstruction of dynamic scenes was explored using Nonrigid Structure-from-Motion~\cite{torresani2008nrsfm}. 

Multi-view stereo methods typically treat dynamic objects as outliers and produce either empty or spurious estimates in moving regions. NSfM methods produce estimates for moving regions 

by solving a reduced dimension approximation of the problem, or by applying additional losses such as smooth-motion priors (see \cite{jensen2020benchmark} for a recent survey).

The effectiveness of an explicit smooth-motion prior depends heavily on accurate tracking, a notoriously difficult problem in itself. NSfM methods use sparse feature tracks (e.g.~\cite{vo2016spatiotemporal}) or build an explicit geometric model and fit it to 2D observations (e.g.~\cite{torresani2008nrsfm}). 
However, feature tracking can only track sparse points, while explicit geometric models fail to capture real-world sequences with complex motions. Instead, we train an MLP to predict a dense scene flow field at all points in 3D, and apply a smooth-motion prior directly to the outputs of the MLP. A similar method of predicting scene flow was used by~\cite{Niemeyer2019ICCV} to align body scans, however they did not take advantage of the unrolling capability of the MLP.

\paragraph{Concurrent Work}

Concurrent to our work, several methods have been proposed to tackle reconstruction of dynamic scenes captured by a moving camera. Dynamic NeRF methods aim at generalizing the original Neural Radiance Field framework~\cite{mildenhall2020nerf} to dynamic scenes by modeling and estimating the scene motion either using a locally-rigid deformation~\cite{park2020nerfies}, or a scene flow field~\cite{li2020neural}. Their training loss is primarily a reconstruction loss over the RGB video frames. These methods have demonstrated impressive results for synthesizing novel views of a dynamic scene, but they  are currently limited to short (2 seconds) videos~\cite{li2020neural}, or small motions~\cite{park2020nerfies}.  In contrast to NeRF-based methods, our main goal is to estimate general purpose depth maps for long videos with arbitrary camera and object motion. 

Robust CVD~\cite{kopf2020robust} also aims to produce depth maps from monocular video, in addition to estimating camera poses. Their method also applies a pre-trained depth prediction model but does not model the 3D motion in the scene, instead assuming that the depth error in dynamic regions can be removed using a spatially and temporally smooth scaling of the initial depth. In contrast, our method finetunes the weights of the depth prediction network while estimating a dense scene-flow field for each frame, and so can resolve initial depth errors that cannot be resolved by smooth scaling.

\newcommand{\flow}[2]{v_{#1\rightarrow #2}}
\newcommand{\occlusion}[2]{MO_{#1\rightarrow #2}}
\newcommand{\stationary}{MS}
\newcommand{\ego}[2]{E_{#1\rightarrow #2}}
\newcommand{\sceneflow}[2]{S_{#1\rightarrow #2}}
\newcommand{\flowof}[3]{v_{#1\rightarrow #2}(#3)}
\newcommand{\egoof}[3]{E_{#1\rightarrow #2}(#3)}
\newcommand{\worldof}[2]{X_{#1}(#2)}
\newcommand{\projof}[2]{\textit{M}_{#1}(#2)}
\newcommand{\displace}[3]{p_{#1\rightarrow #2}(#3)}
\newcommand{\reproj}[3]{p_{#1\rightarrow #2}(#3)}
\newcommand{\dcnnof}[1]{F_{\theta_d}(#1)}
\newcommand{\sfcnnof}[2]{G_{\theta_s}(#1, #2)}
\newcommand{\dcnn}[0]{F_{\theta_d}}
\newcommand{\sfcnn}[0]{G_{\theta_{s}}}
\newcommand{\dispdiff}[3]{\mathcal{L}^{\mathit{disp}}_{#1\rightarrow #2}(#3)}
\newcommand{\screendiff}[3]{\mathcal{L}^{\mathit{2D}}_{#1\rightarrow #2}(#3)}
\newcommand{\prior}{\mathcal{L}^{\mathit{prior}}}
\newcommand{\staticprior}{\mathcal{L}^{\mathit{static}}}
\newcommand{\disploss}{\mathcal{L}^{\mathit{disp}}}
\newcommand{\flowloss}{\mathcal{L}^{\mathit{2D}}}
\newcommand{\totalloss}{\mathcal{L}^{\mathit{total}}}
\newcommand{\reprojdepthof}[3]{D_{#1\rightarrow#2}(#3)}
\newcommand{\absnorm}[1]{\left\lVert #1 \right\rVert_1}

\section{Method}

\subsection{Overview}
\label{sec:overview}
\begin{figure}[t!]
    \centering
    \includegraphics[width=\columnwidth]{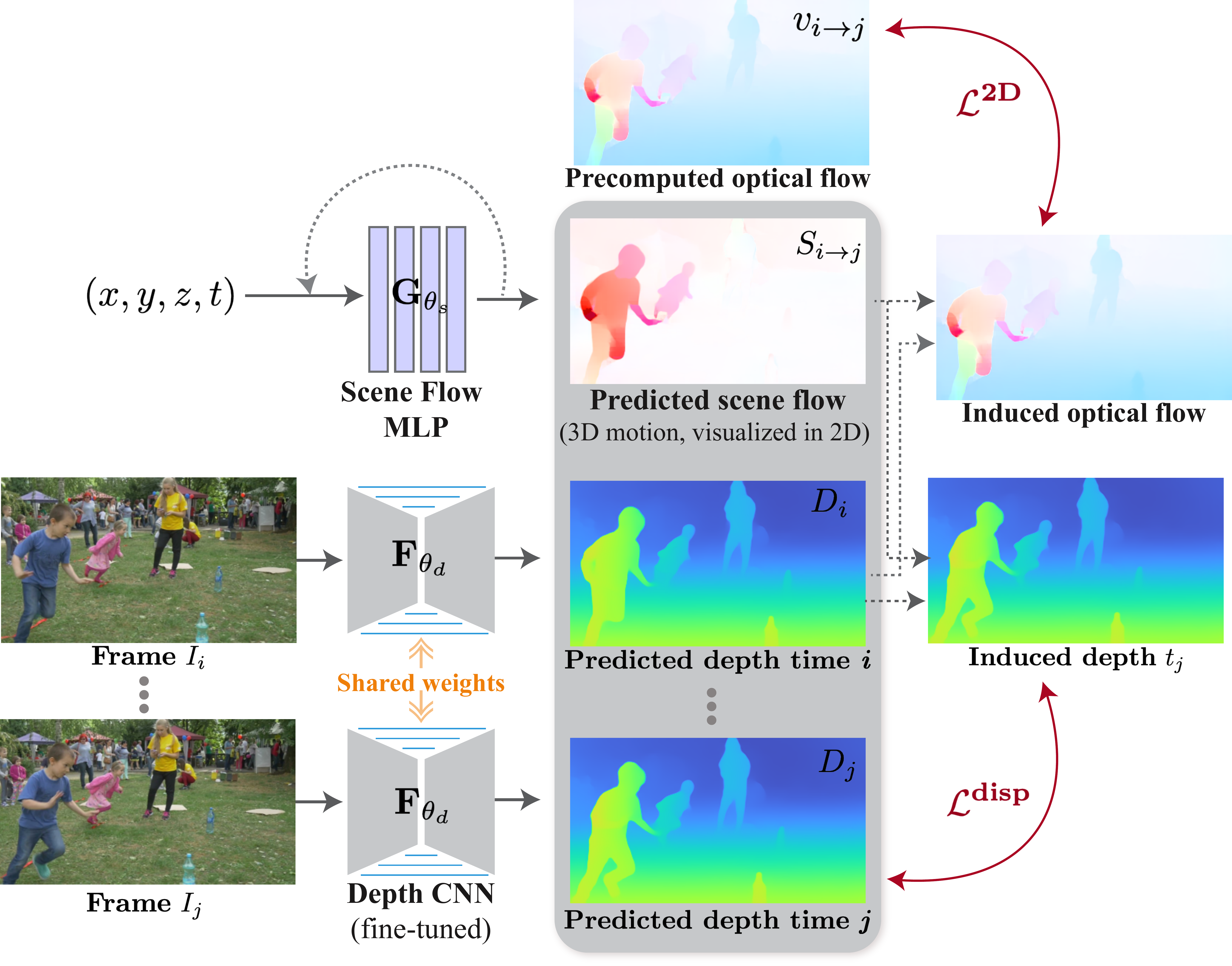}
    \caption{{\bf Pipeline.} Our system takes as input an ordinary video with freely-moving camera and scene objects and predicts a depth map for each frame of the video. We assume known camera poses and compute optical flow between frames in a per-processing step. 
    Our system consists of two networks: A \emph{depth prediction CNN} $F_{\theta_d}$, which takes $I_i$ (RGB frame) as input and outputs a depth map $D_i$; we initialize this network with a pre-trained monocular depth model (e.g., \cite{li2020mannequinchallenge}); and  a \emph{scene flow MLP} $G_{\theta_s}$, which takes a space-time point, $(x,y,z,t)$, as input and outputs its scene flow vector, i.e., 3D displacement vector w.r.t. $t+1$. The depth and scene flow networks are trained jointly in a self-supervised manner under two losses: $\flowloss$ measures the difference between the induced flow and the input optical flow, while $\disploss$ measures the difference between the induced depth at time $j$ and the predicted depth at time $j$. We further apply a constant velocity prior on the scene flow vectors (see Section~\protect\ref{sec:losses}).
    }
    \label{fig:system_diagram}
\end{figure}

 Figure~\ref{fig:system_diagram} illustrates the pipeline of our method. Our system takes a calibrated video as input where both the objects in the scene and the camera are naturally moving, and predicts per-frame depth maps. This is done via an \emph{optimization framework} where two networks are trained, at test time, on the input video: (i) a \emph{single-frame CNN depth prediction model} that takes an RGB image as input and outputs a depth map. (ii) a \emph{scene flow MLP prediction model} that takes a 4D point $(x,y,z,t)$ as input and outputs its 3D displacement to the next time step $t+1$. Both the depth and the scene flow networks are \emph{trained jointly} where the depth network is initialized with a pre-trained monocular depth prediction model (e.g., \cite{li2020mannequinchallenge} or \cite{Ranftl2020}), and then finetuned over the input video; this initialization equips our model with depth priors learned from external source of data~\cite{Luo-VideoDepth-2020}. The scene flow module allows us to explicitly model the 3D motion of arbitrary dynamic objects in the scene, and is trained from scratch.

  In a pre-processing step, we compute camera poses of all the frames and optical flow fields between pairs of frames $\{I_i, I_j\}$ using off-the-shelf techniques (Section~\ref{sec:inputs}). The computed camera poses and optical flow fields are used to apply two types of re-projection losses: a \emph{flow consistency loss}, i.e., we require that the  2D displacement field between two frames resulting by  projecting the predicted depth and scene flow onto 2D would match the pre-computed 2D optical flow field between the frames; we further apply a \emph{depth consistency loss} which encourages coherency between the depth and scene flow estimates across different frames.  Finally, we impose a local linear motion prior on the predicted scene flow to reduce the inherent ambiguity between depth and motion. We next describe each of these losses in detail.

\begin{figure}
    \centering
    \includegraphics[width=\columnwidth]{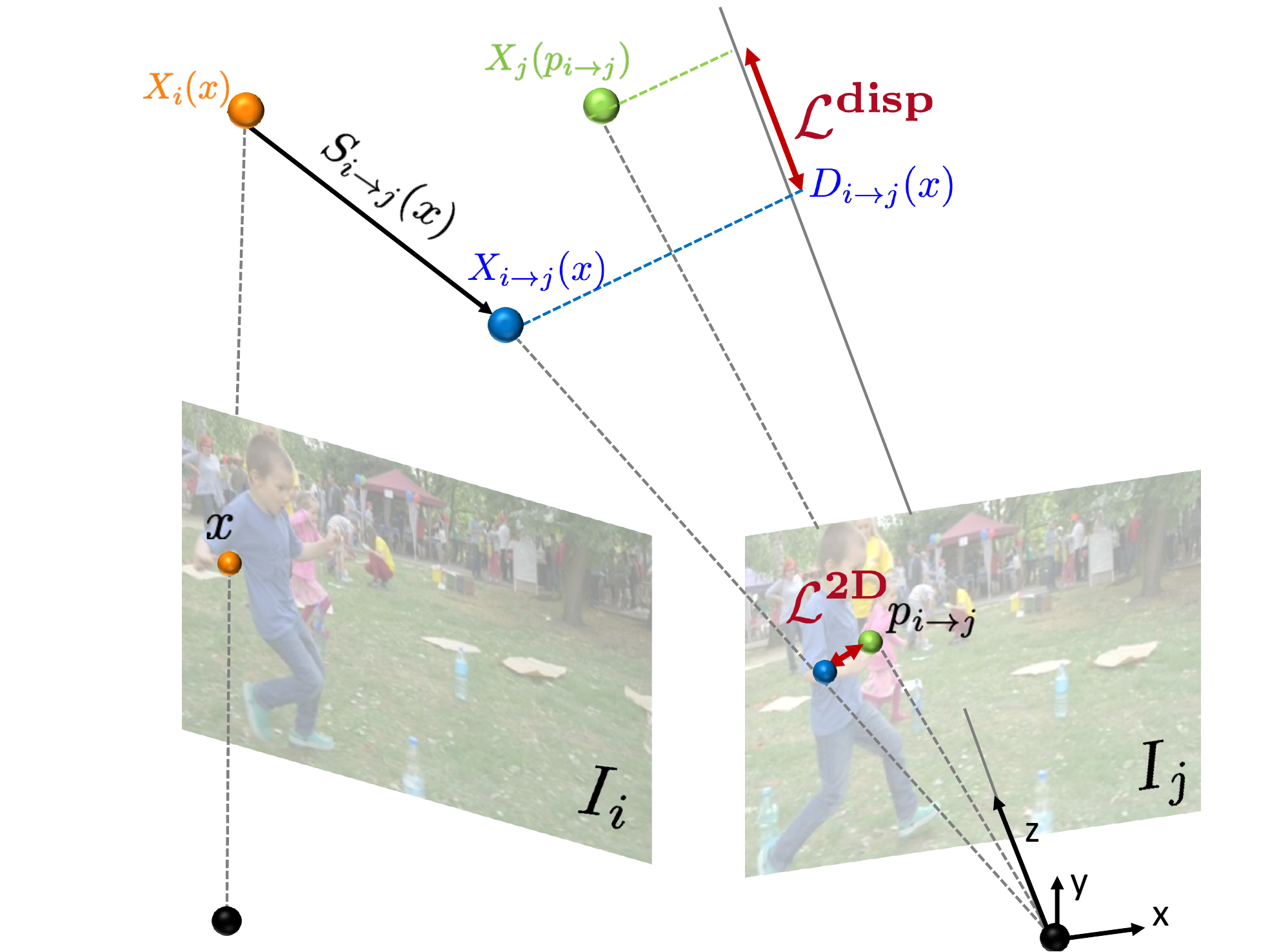}
    \caption{{\bf Depth and scene flow training losses.} A pixel $x$ in frame $i$ is backprojected into a 3D point $X_i(x)$ (orange point), using our predicted depth map value $D_i(x)$. Similarly, the corresponding pixel of $x$ in frame $j$, denoted by $p_{i\rightarrow j}$, is backprojected to $X_j(p_{i\rightarrow j})$ (green point). We require consistency between $X_j(p_{i\rightarrow j})$ and $X_i(x)$ after displacement by our predicted scene flow $S_{i\rightarrow j}$, denoted by $X_{i \rightarrow j}(x)$ as follows: (i) $\mathcal{L}^\textbf{2D}$ penalizes the 2D distance between the projection of $X_{i \rightarrow j}(x)$ onto frame $j$, and $p_{i\rightarrow j}$. (ii) $\mathcal{L}^\textbf{disp}$ penalizes the difference in disparity between $X_{i \rightarrow j}(x)$ and $X_j(p_{i\rightarrow j})$. See details in Section.~\ref{sec:losses}. }
    \label{fig:losses}
\end{figure}

\subsection{Pre-processing}
\label{sec:inputs}

We use a similar pre-processing strategy by presented by ~\cite{Luo-VideoDepth-2020}. 
The input RGB frames $I_i$ are first triangulated with ORB-SLAM2~\cite{murTRO2015} for initial pose estimates, which is later refined by the structure-from-motion stage of COLMAP \cite{schoenberger2016sfm} to produce camera poses $R_i$, $t_i$ and sparse depth maps $D_i^\mathit{sfm}$ for every frame. For sequences where motion masks are available, we use the multi-view stereo stage of COLMAP~\cite{schoenberger2016mvs} for more accurate poses and denser depth maps. Then, forward optical flow $\flow{i}{i+k}$ and backwards optical flow $\flow{i+k}{i}$ are computed between subsequent frames ($k=1$) and between a subset of wide-baseline frame pairs ($k \in [2,4,6,8]$) using RAFT~\cite{teed2020raft}. 
We then generate initial depth maps $D_i^\mathit{init}$ using a single-frame depth prediction network~\cite{li2020mannequinchallenge,Ranftl2020}. Since such predictions are scale-invariant, we align the scale of $t_i$ to roughly match the the scale of initial depth estimates. Specifically, the scale $s$ is calculated by: $s = \mathit{mean}(\mathit{median}(D_i^\mathit{init}/D_i^\mathit{sfm}))$, and applied for all camera translations $t_i$.

For each pair of optical flow fields $\flow{i}{i+k}$ and $\flow{i+k}{i}$ we find occluded regions (as well as regions of inaccurate flow) using forward-backward consistency to generate an occlusion mask $\occlusion{i}{i+k}$ for masking the loss computation. All flows with inconsistency larger than $1$ pixel are considered to be occluded or unreliable. Formally: 
  \begin{equation}
    \occlusion{i}{i+k}(x)=
    \begin{cases}
      1, & \text{if}\ \left\lvert \flow{i}{i+k}(x) + \flow{i+k}{i}(x+\flow{i}{i+k}) \right\rvert_2 > 1\\
      0, & \text{otherwise}
    \end{cases}
  \end{equation}
where $x$ denotes pixel locations in frame $i$. 

\subsection{Test-time Learning of Depth and Scene Flow}
\label{sec:losses}

Given the input video, along with the pre-computed optical-flow fields and camera poses (see Section~\ref{sec:inputs}), we turn to the task of fine-tuning a pre-trained, single-frame, depth-prediction network, $\dcnn$ that takes an RGB frame $I_i$ as input and produces a depth map $D_i$, and training from scratch an auxiliary scene-flow-prediction network $\sfcnn$ that takes a 3D point in world coordinate system $X$, and predicts its 3D world-space scene flow $\sceneflow{i}{i+1}$. 

The networks' parameters $\theta_d$ and $\theta_s$ are the only variables of the optimization and are trained in conjunction: the inputs to the scene flow network are updated at each optimization step, based on the current depth estimate; and the scene flow losses are back-propagated to update our depth model at each step.

Theoretically, the scene flow module is redundant---it can be analytically computed given 2D dense correspondences between frames (optical flow) and depth estimates for each frame. However, such estimation can often be noisy and unstable especially when the camera rays at corresponding points are nearly parallel.  Implicitly representing the scene flow using our MLP network acts as an auxiliary variable that is encouraged to match the analytically computed scene flow through the training losses. This design choice is crucial for making the optimization stable as demonstrated in Section.~\ref{sec:cube_analysis}.

Our objective loss for optimizing $\sfcnn$ and $\dcnn$ consists of three terms:
\begin{equation}
 \argmin_{\theta_d, \theta_s} \flowloss + \alpha \disploss + \beta \prior 
 \label{eq:total_loss}
\end{equation}
 $\flowloss$ encourages the depth and scene flow between two frames to match the pre-computed optical flow when projected onto 2D. $\disploss$ encourages  the predicted depth $D_i$ and scene flow w.r.t. frame $j$, $\sceneflow{i}{j}$ to be consistent with $D_j$, the predicted depth of frame $j$. $\prior$ is a smoothness prior imposed on the predicted scene flow. 
 The hyper-parameters $\alpha$ and $\beta$ control the relative weighting of the different terms. We next describe each of the loss terms in detail. 

\subsection{Training Losses}
\label{sec:reprojection_loss}
Given a pair of frames $\{I_i, I_j\}$, we define the following losses on the predicted depth maps $\{D_i, D_j\}$, and the predicted scene flow between the frames $\sceneflow{i}{j}$. An visual illustration of these losses and our notation is presented in Fig.~\ref{fig:losses}.

We unproject each pixel $x\in I_i$ into a 3D point $X_i(x)$ in world coordinates, using the predicted depth map $D_i$ and camera poses:

\begin{equation}
\worldof{i}{x} = R_i(D_i(x) K_i^{-1} \tilde{x}) + t_i, 
\label{eq:unproject}
\end{equation}
where $\tilde{x}$ is the 2D homogenous augmentation of $x$, $K_i$ is the camera intrinsics matrix, $R_i$ is the 3x3 camera rotation matrix, and $t_i$ is the camera translation vector.

We can now compute the scene flow of $X_i(x)$ w.r.t. time $i+1$ by feeding it into our MLP scene flow network. That is, 
\begin{equation}
    S_{i\rightarrow i+1 }(x) = \sfcnnof{X_i(x)}{i}.
\end{equation} 
To compute the scene flow w.r.t. frame $j$ where $j-i > 1$, we can simply unroll the scene flow MLP $j-i$ times:
\begin{equation}
\begin{array}{l}
 S_{i\rightarrow j }(x) = \sfcnnof{\tilde{X}_{j-1}}{j-1}, \vspace{0.1cm} \\ 
 \tilde{X}_k = \tilde{X}_{k-1} + S_{k-1\rightarrow k }(x), \qquad k=i+1, ..., j-1.
 \end{array}
 \label{eq:unroll}
\end{equation}
Finally, the 3D position of $X_i(x)$ at time $j$ is given by:
\begin{equation}
    X_{i \rightarrow j}(x) = X_i(x) + \sceneflow{i}{j}(x).
\end{equation}

\paragraph{ Consistency in 2D} We define a 2D consistency loss $\screendiff{i}{j}{x}$ that measures the  pixel distance between the corresponding pixel of $x$ in frame $j$, and the projection of $X_{i \rightarrow j}(x)$ onto camera $j$.

Formally, let $\flowof{i}{j}{x}$ be the pre-computed optical flow vector of pixel $x\in I_i$  w.r.t. frame $I_j$. The corresponding pixel of $x$ is given by:
\begin{equation}
    \displace{i}{j}{x} = x + \flowof{i}{j}{x}.
\end{equation}
The loss term $\flowloss$ is given by:
\begin{equation}
\begin{array}{l}
    \screendiff{i}{j}{x} = \absnorm{\projof{j}{\worldof{i \rightarrow j}{x}} - \displace{i}{j}{x}}, \vspace{0.15cm} \\
    \flowloss = \sum_{i,j}\sum_{x\in \{\occlusion{i}{j}=0\}}\screendiff{i}{j}{x},
\end{array}
\end{equation}
where $\absnorm{\cdot}$ denotes the $L_1$ norm, $\occlusion{i}{j}$ denotes the occlusion mask of frame $i$ w.r.t frame $j$,  and $\projof{j}{X}$ denotes the projection of a world-space point $X$ onto the $j^{th}$ camera:
\begin{equation}
 \projof{j}{X_i} = \pi(K_j R^T_j (X_i - t_j)),
\end{equation}
where $\pi$ is the projection operator $\pi([x,y,w]^T) = [x/w,y/w]^T$.

\paragraph{Consistency in 3D} 
Similarly to~\cite{Luo-VideoDepth-2020}, we define the disparity consistency loss $\disploss$, which measures the consistency between the inverse depth value of $\worldof{i \rightarrow j}{x}$ under camera $j$, and the inverse of $D_j$:
\begin{equation}
\begin{array}{l}
    \reprojdepthof{i}{j}{x}  = |K_j R^T_j (X_i +\sceneflow{i}{j}(x)- t_j)|_z  \vspace{0.15cm}\\
    \dispdiff{i}{j}{x}  = \absnorm{\frac{1}{\reprojdepthof{i}{j}{x}} - \frac{1}{D_j(\displace{i}{j}{x})}} \vspace{0.15cm}\\
    \disploss = \sum_{i,j}\sum_{x\in \{\occlusion{i}{j}=0\}}\dispdiff{i}{j}{x}
\end{array}
\end{equation}
where $|\cdot|_z$ denotes taking the depth component of a 3D point, i.e. $z = |[x,y,z]|_z$. 

\paragraph{ Smooth 3D motion} To regularize the motion in 3D, we impose a constant velocity prior $\prior$:
\begin{equation}
    \begin{array}{c}
    \prior_i(x) = \absnorm{\sceneflow{i}{i+1}(x) - \sfcnnof{\worldof{i}{x}+\sceneflow{i}{i+1}(x)}{i+1}}, \vspace{0.15cm} \\
    \prior = \sum_{i}\sum_{x\in I_i}\prior_i(x)
    \end{array}
    \label{eq:motion_prior}
\end{equation}
where $x$ is the pixel location in frame $i$.

\paragraph{ Optional motion masks} If motion segmentation is available for the video sequence, we can further constrain the motion in 3D by penalizing velocities in static regions. Formally, given a mask of static regions at frame $i$, denoted as $\stationary_i$, we introduce a static prior $\staticprior$:
\begin{equation}
    \begin{array}{l}
    \staticprior_i(x) = \absnorm{\sceneflow{i}{i+1}(x)}, \vspace{0.2cm} \\
    \staticprior = \sum_{i}\sum_{x\in \stationary_i}\staticprior_i(x).
    \end{array}
\end{equation}
The total loss when using static/dynamic masks is given by:

\begin{equation}
    \totalloss_\mathit{mask} = \flowloss +  \alpha\disploss +\beta\prior+\gamma\staticprior.
\label{eq:total_loss_motion_seg}
\end{equation}
Otherwise, the loss is given by Eq.~\ref{eq:total_loss}.
We \textbf{do not} use any motion segmentation throughout our experiments unless explicitly noted; for real videos, only the ablation study (Fig.~\ref{fig:motion_mask}) includes a motion mask.

\paragraph{ Training schedule}
\label{sec:training_schedule}
As discussed in Section~\ref{sec:overview}, we initialize the depth network with a pre-trained monocular depth model. However, the scene flow network $\sfcnn$ has to be trained  from scratch. Naively initializing $\sfcnn$ with random weights produces small initial scene flow values that are inconsistent with the initial depth estimate and pre-computed optical flow. Training both depth and scene flow networks from this initialization  results in large changes in the depth prediction network in order to satisfy the consistency losses. We overcome this issue via a ``warm-up" phase in which only $\sfcnn$ is trained using $\totalloss$ with $\beta=0$ while the pre-trained depth network $\dcnn$ is kept fixed. In practice, we found $\sfcnn$ converged after $5$ epochs, and set the warm-up phase to $5$ epochs in all our experiments, after which both networks are finetuned jointly using $\totalloss$.

Note that even if the predicted scene flow matches the analytic scene flow perfectly after warm-up, there is still a training signal from the smooth-motion prior applied to the scene flow after the warm up phase. The scene flow MLP will deviate from its warmed-up state during training in order to find a solution that induces locally linear motion while satisfying the optical flow loss terms.

\section{Implementation Details}

\paragraph{ Network architectures}
\label{sec:network_arch}
In our experiments , we use either the single-frame MannequinChallenge model~\cite{li2019learning} or the MiDaS v2 model~\cite{Ranftl2020} for the depth prediction network $\dcnn$. The scene flow network $\sfcnn$ is a MLP with positional encoding~\cite{mildenhall2020nerf} for the input. Specifically, the positional encoding $\xi(x)$ for input $x$ is defined as:
$$\xi(x) = [\sin{\pi x}, \cos{\pi x}, \sin{2\pi x}, \cos{2\pi x}, ...\sin{N\pi x}, \cos{N\pi x}], $$
where we use $N=16$ for all the experiments, and the encoding is applied to both 3D locations and time. The MLP is comprised of $4$ hidden layers, each with $256$ units. 

\paragraph{Hyperparameters} Throughout our experiments, we set $\alpha=0.1$, $\beta=1$ for the total loss defined in Eq.~\ref{eq:total_loss}. When motion segmentations are present, we set $\gamma=100$ in Eq.~\ref{eq:total_loss_motion_seg}. We use a learning rate of $10^{-5}$ for the depth network when initialized with the MannequinChallenge model~\cite{li2019learning}, and a learning rate of $10^{-6}$ when initialized with the MiDaS v2 model ~\cite{Ranftl2020}. The sceneflow network is optimized with a learning rate of $0.001$. Both networks are optimized using the Adam optimizer with $\beta_1=0.9$ and $\beta_2=0.999$. Batch size is set to $1$.

\paragraph{ Training Time} The networks are trained for 20 epochs for all experiments, so the training time of our method depends linearly on the number of frames. For a 60-frame video, the training time is about 75 minutes on a single NVIDIA P100 GPU. The training time of our method is about 2$\times$ longer than CVD~\cite{Luo-VideoDepth-2020}, due to the extra scene flow network.

\section{Results}
\label{sec:results}
Our experiments demonstrate the quality of predicted depth and the importance of the scene flow network on two synthetic datasets and real-world videos. We first justify the design choice of the scene flow MLP against analytically deriving scene flow (Sec.~\ref{sec:cube_analysis}). We then evaluate our method on the synthetic Sintel~\cite{sintel} dataset and show quantitative improvement over competing methods (Sec.~\ref{sec:sintel_score}). To demonstrate the real-world capabilities of the method, we show qualitative examples on videos with significant camera and object motion (Sec.~\ref{sec:qualitative}). Finally, we show multiple video editing results based on our consistent depth estimates (Sec.~\ref{sec:applications}).

\subsection{MLP scene flow vs. Analytic scene flow}
\label{sec:cube_analysis}
\begin{figure}[t!]
    \centering
    \includegraphics[width=\columnwidth]{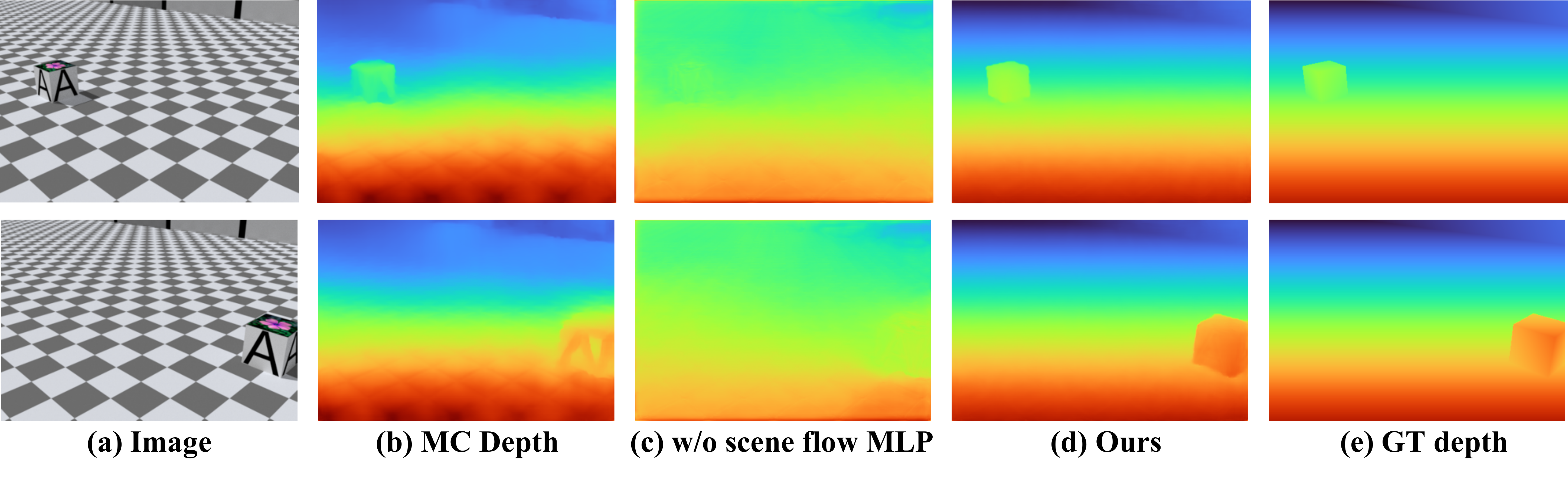}
    \caption{{\bf MLP scene flow vs. analytic scene flow.} (a) Sample frames from the \emph{cube} video -- the cube is moving along a straight line in 3D, while the camera is moving side-to-side. (b) Initial depth maps computed by MC~\protect\cite{li2019learning}. (c) Results produced when only the depth network is optimized, while the scene flow is computed analytically ('w/o scene flow MLP'); this baseline fails to converge to the correct solution  (e). Our method (d) produces near perfect results under this setup. In both (c) and (d), we use ground truth optical flow and camera poses. See more details in Section~\ref{sec:cube_analysis}.}
    \label{fig:cube_compare}
\end{figure}

We examine the importance of representing scene flow implicitly via our MLP model vs. analytically computing scene flow (using the depth estimates, camera poses and optical flow fields). To do so, we generated a simple synthetic scene where a cube is moving along a straight line in 3D, towards the camera at a constant velocity, while the camera is moving along a sine curve that simulates hand-held camera wobble (see Figure~\ref{fig:cube_compare}a, and full video in the SM). Using ground truth optical flow fields and camera poses, we then train:
\begin{enumerate}
    \item Our full framework, including both depth and scene flow networks, using Eq.~\ref{eq:total_loss}.
    \item Only depth network only (no MLP scene flow). At each iteration of training, we compute scene flow given the current depth estimates:
    $$
\hat{S}_{i\rightarrow i+1}(x) = X_{i+1}(p_{i\rightarrow i+1}(x)) - X_i(x).
$$
where $X_i(x)$ and $X_{i+1}(p_{i\rightarrow i+1}(x))$ are computed by unprojecting pixel $x\in I_i$, and its corresponding pixel $p_{i\rightarrow i+1}(x)\in I_{i+1}$ to 3D, using the depth estimates $\{D_i, D_{i+1}\}$ as defined in Eq.~\ref{eq:unproject}. We then plug the computed scene flow into Eq.~\ref{eq:total_loss} by replacing all $S_{i\rightarrow i+1}$ and $\sfcnnof{X_i}{i}$ with $\hat{S}_{i\rightarrow i+1}$.  Note that by construction,  $\flowloss$ and $\disploss$ are perfectly satisfied (the scene flow is derived from two depth maps and flow). Thus, the only loss that drives the optimization is $\prior$. Intuitively, this baseline simply optimizes the depth network such that the derived scene flow is locally smooth.
\end{enumerate}

As can be seen in Figure~\ref{fig:cube_compare}, the baseline w/o scene flow MLP does not converge to the correct solution, even when ground truth optical flow and camera poses are used to derive scene flow. 

We hypothesize the failure is due to the $\prior$ term being ill-posed for analytic scene flow. When the motion of the camera and the object are nearly aligned, even momentarily, $p_{i\rightarrow i+1}(x) \approx x$. In this case analytic scene flow is ill-posed (see Appendix~\ref{appendix:analytic_sf}) and the gradient of $\prior$ is numerically unstable, leading to poor convergence.

In contrast, our scene flow network $\sfcnn$ avoids this issue by acting as a slack variable that aggregates information across the entire video. When $\prior$ is defined in terms of the output of $\sfcnn$ (Eq.~\ref{eq:motion_prior}), the scene flow at $\worldof{i}{x}$ potentially depends on \emph{all the rays in the scene}.
Each frame of the video has $\approx 2^{17}$ pixels, whereas $\sfcnn$ has only $\approx 2^{18}$ parameters, so weights must be shared between rays.

In the presence of badly conditioned ray pairs these shared weights act as a regularizer, allowing our method to nearly perfectly reconstruct the ground truth cube (Fig.~\ref{fig:cube_compare}d).

\subsection{Quantitative Evaluation}
\label{sec:sintel_score}
\begin{table*}[ht]
\caption{\textbf{Depth accuracy on Sintel.}  
We evaluate error for the entire depth map (Full), and separately for the dynamic and static regions.  We evaluate the initial depth maps produced by MC (top) and MiDaS (bottom),  along with the depth prediction results of CVD and our method using each of the depth initializations. Additionally, we evaluate our performance when the smooth motion prior is omitted from our objective (w/o $\prior$), and when static/dynamic segmentation masks are incorporated (Eq.~\ref{eq:total_loss_motion_seg}). Our full method consistently improves upon the initial depth estimations, in both static and moving regions. Since CVD assumes a static scene, it fails to recover large moving regions. Given the motion masks, our performance further improves in static and dynamic regions, and is on par with CVD in static regions. Additionally, we observe a significant degradation in our performance when  $\prior$ is omitted, demonstrating its importance in reducing the ambiguity between depth and 3D motion. CVD$^\dagger$ denotes our implementation of CVD under MiDaS initialization, as the original code fails to train on some of the sequences (see Sec.~\ref{sec:results} for more details.)} 
\centering
\begin{tabular}{l|ccc|ccc|ccc}
\toprule
Method & \multicolumn{3}{c}{Full} & \multicolumn{3}{c}{Dynamic} & \multicolumn{3}{c}{Static}\\
 & $L_1$ rel. $\downarrow$  & log RMSE $\downarrow$ & RMSE $\downarrow$ & $L_1$ rel. $\downarrow$  & log RMSE $\downarrow$ & RMSE $\downarrow$ &  $L_1$ rel. $\downarrow$  & log RMSE $\downarrow$ & RMSE $\downarrow$ \\
\midrule

MC~\cite{li2019learning} & 0.6508 & 1.1978 & 4.0830 & 0.5083 & 0.4974 & 3.6870 & 1.3691 & 1.3443 & 2.9392\\
CVD, MC init~\cite{Luo-VideoDepth-2020} & 1.5208 & 0.6874 & 3.3531 & 2.9116 & 0.8814 & 4.9374 & \textbf{0.3291} & \underline{0.2902} & \underline{1.8832}\\
Ours, MC init, w/o $\prior$ & 0.6753 & 0.7837 & 3.7396 & 0.8227 & 0.6575 & 3.7171 & 0.6654 & 0.6803 & 2.5594 \\
Ours, MC init & \underline{0.5280} & \underline{0.6907} & \underline{3.1284} & \underline{0.4841} & \underline{0.5655} & \underline{3.0244} & 0.9963 & 0.5723 & 2.2443\\
Ours, MC init, w/ mask & \textbf{0.3540} & \textbf{0.4877} & \textbf{2.8530} & \textbf{0.4707} & \textbf{0.5464} & \textbf{2.8469} & \underline{0.3494} & \textbf{0.2845} & \textbf{1.8480}\\
\midrule
MiDaS~\cite{Ranftl2020}  & 0.4506 & 0.7440 & 3.2763 & 0.5202 & 0.6361 & 2.5613 & 0.5743 & 0.6243 & 2.7371\\ 
$\text{CVD}^\dagger$, MiDas init & 0.6245 & 1.0754 & 4.7060 & 1.7494 & 1.4744 & 6.9751 & \textbf{0.2321} & \textbf{0.3154} & \underline{1.6662}\\
Ours, MiDas init, w/o $\prior$ & 0.4095 & 1.7713 & 3.8590 & 0.5758 & 0.6400 & 2.6002 & 0.2865 & 0.5438 & 2.7245\\
Ours, MiDas init & \underline{0.3838} & \underline{0.5097} & \underline{2.6733} & \underline{0.4520} & \underline{0.4997} & \underline{2.5374} & 0.3746 & 0.3814 & 1.9427\\
Ours, MiDas init, w/ mask & \textbf{0.3063} & \textbf{0.4708} & \textbf{2.4941} & \textbf{0.4468} & \textbf{0.3990} &  \textbf{2.5018} & \underline{0.2455} & \underline{0.3347} & \textbf{1.6082}\\
\bottomrule
\end{tabular}
\vspace{0.3em}
\label{tab:accuracy}
\end{table*}

\begin{figure*}[ht!]
    \centering
    \includegraphics[width=\linewidth]{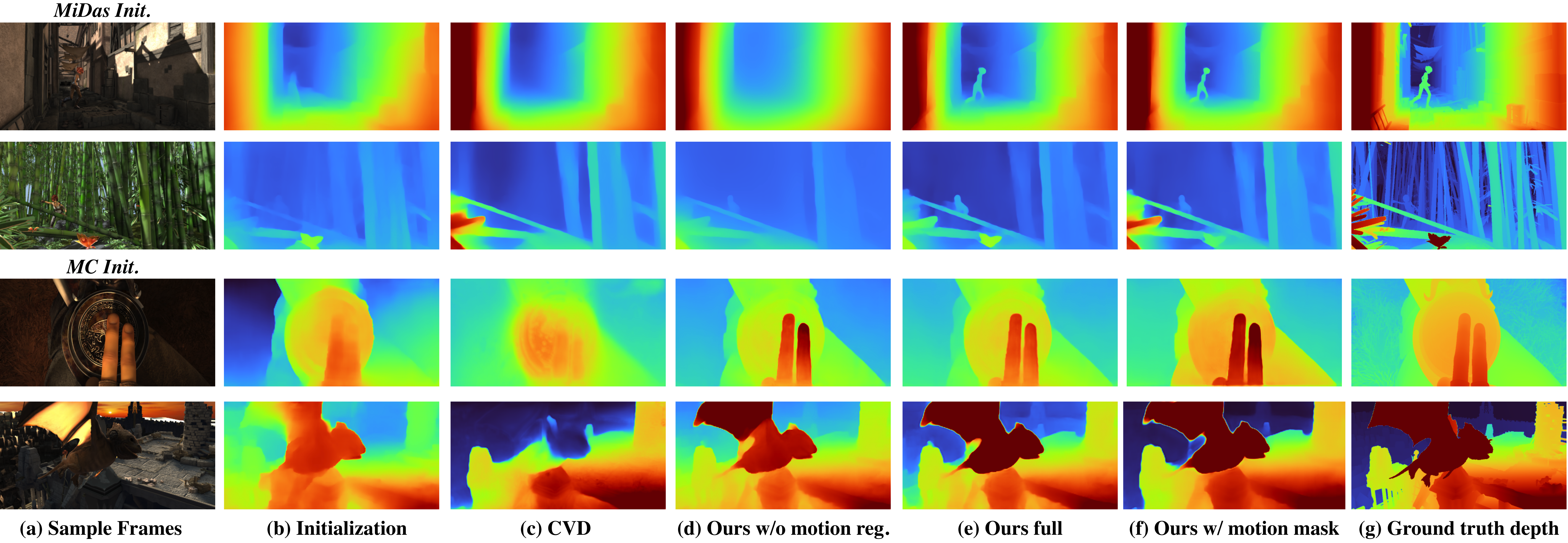}
    \caption{ \textbf{Qualitative results on Sintel.} Variants of our method are shown in (d,e,f) using MiDaS v2~\mbox{\cite{Ranftl2020} as initialization}, along with initial depth estimates (b) and CVD (c). Top two rows show initialization with MiDaS, while bottom two show initialization with MC. Both MiDaS and MC initialization are roughly correct but show visible errors. CVD improves static regions over the initialization, but fails on moving regions. Without the acceleration regularizer, our method fails to recover the moving character (d, top row); with the regularizer (e), the character is put at the correct depth. Adding a motion mask to our method (f) further improves the depth prediction quality on static regions.\vspace{-5pt}}
    \label{fig:sintel_results}
\end{figure*}

\begin{figure*}[ht!]
    \centering
    \includegraphics[width=\textwidth]{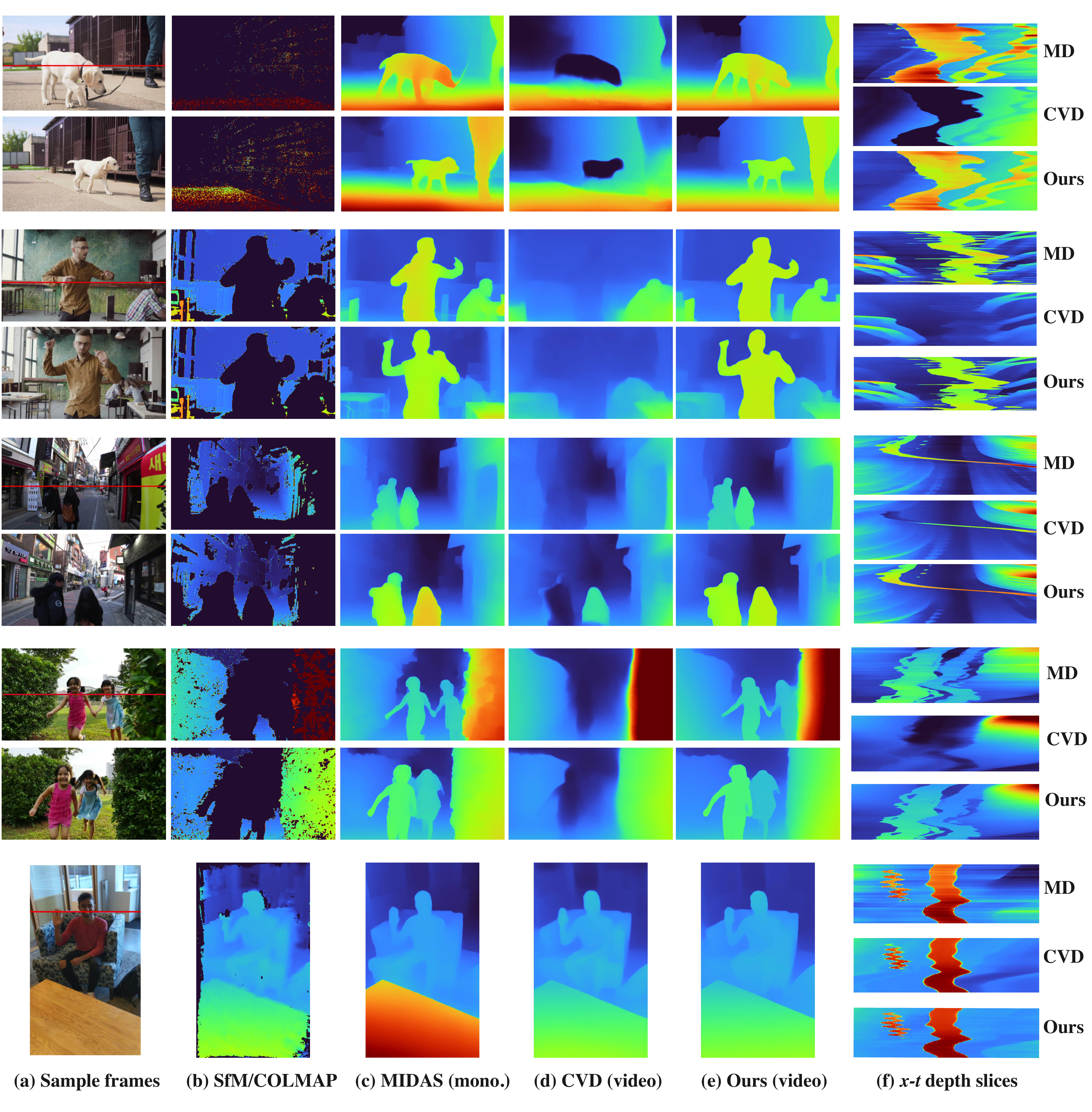}
    \caption{{\bf Results on real videos.} (a) Two representative video frames. (b) Sparse depth maps produced by running SfM/MVS (see Section~\ref{sec:inputs});  top two rows show  COLMAP SfM output, and the rest COLMAP MVS results (dynamic regions are filtered out and assigned zero depth). (c)
     Initial depth maps produced by MiDaS (single-frame model), and by (d) CVD which optimizes single-frame depth prediction model over the input video. (e) Our predicted depth maps (no motion segmentation masks were used). (f) Corresponding $x-t$ slice for each of the methods (the sampled horizontal line is marked in red in (a)): MiDas produces inconsistent depth over time, apparent by the zigzagging patterns in the slice. CVD  outputs temporally consistent depth estimates in the static regions, but fails in the dynamic regions (darker regions in the slice). Note that the last row depicts small, in-place motion (person is waving), in which case most of the person region is reconstructed by MVS. In all example, our method is able to produce temporally consistent depth estimates in both static and dynamic regions.}
    \label{fig:res_real}
\end{figure*}

We quantitatively evaluate, ablate  and compare our method on a subset of the Sintel dataset ~\cite{sintel}, as follows.

\paragraph{ Dataset} 
We test the depth prediction accuracy of our method using the 23 sequences of the Sintel dataset. We use the motion masks provided by \cite{Taniai2017} as $MS_i$ and for separately evaluating performance in static and dynamic regions. Input optical flow is computed using RAFT~\cite{teed2020raft}, not the ground-truth flow from the dataset.

\paragraph{ Baselines} We compare our method with three baselines: the single-frame MannequinChallenge model(MC)~\cite{li2019learning}, single-frame MiDaS v2~\cite{Ranftl2020} and CVD. For CVD with MC initialization, we use the original implementation provided by the authors. We found training the original implementation of CVD numerically unstable (training loss being NaN for some sequences)  when initialized with MiDaS due to the near-zero disparity values MiDaS produces in background areas. We therefore implemented a modified version of CVD that clips near-zero disparity to a small, non-zero value using our frame work: removing the scene flow MLP and replace our losses with the ones proposed by the original paper. 

\paragraph{ Ablation.} We compare our method when:   the smooth motion prior is omitted from our objective (``w/o motion prior''), and w/o and w/ dynamic/static motion segmentation masks (Eq.~\ref{eq:total_loss} and Eq.~\ref{eq:total_loss_motion_seg}, respectively.) 

\paragraph{ Metrics} We use three metrics to evaluate the quality of the depth maps: $L_1$ relative, log RMSE, and RMSE~\cite{sintel}. To examine temporal consistency, we do not match the scale of each individual frame against the ground truth during evaluation. Instead, we apply a single calibrated scale for the entire sequence during the preprocessing stage, as described in Sec.~\ref{sec:inputs}. We also evaluate separately on moving and static regions using the ground truth motion mask. We follow the common practice of excluding depth values that exceed 80m during evaluation, as distant, high-magnitude outliers otherwise confuse the results. 

As can be shown in Table~\ref{tab:accuracy}, our method, when initialized with the corresponding single-frame depth prediction model, consistently improves upon the initialization in both moving and static regions by an average of 40-45\% in $L_1$ relative error. Our method outperforms CVD in moving regions by a large margin due to our explicit handling of motion, while CVD still performs well in static regions. When motion masks are incorporated into our framework,  we achieve similar performance to CVD in static regions. Finally, removing the motion prior $\prior$ for our objective leads to 150-175\% increase in $L_1$ relative error, which demonstrates the importance of this regularization. 

Figure~\ref{fig:sintel_results} shows several qualitative results. Note that our method (e) using MiDaS v2~\cite{Ranftl2020} as initialization improves depth in both moving and static regions compare to the initial depth estimates (b) even when motion segmentation masks are not used. Using the motion masks further improves our results (e,f).  CVD improves the static regions but fails to produce correct depth for moving objects, tending to erase them to the background value (c).

\subsection{Real Videos -- Qualitative Results}
\label{sec:qualitative}
\begin{figure}[t!]
    \centering
    \includegraphics[width=\columnwidth]{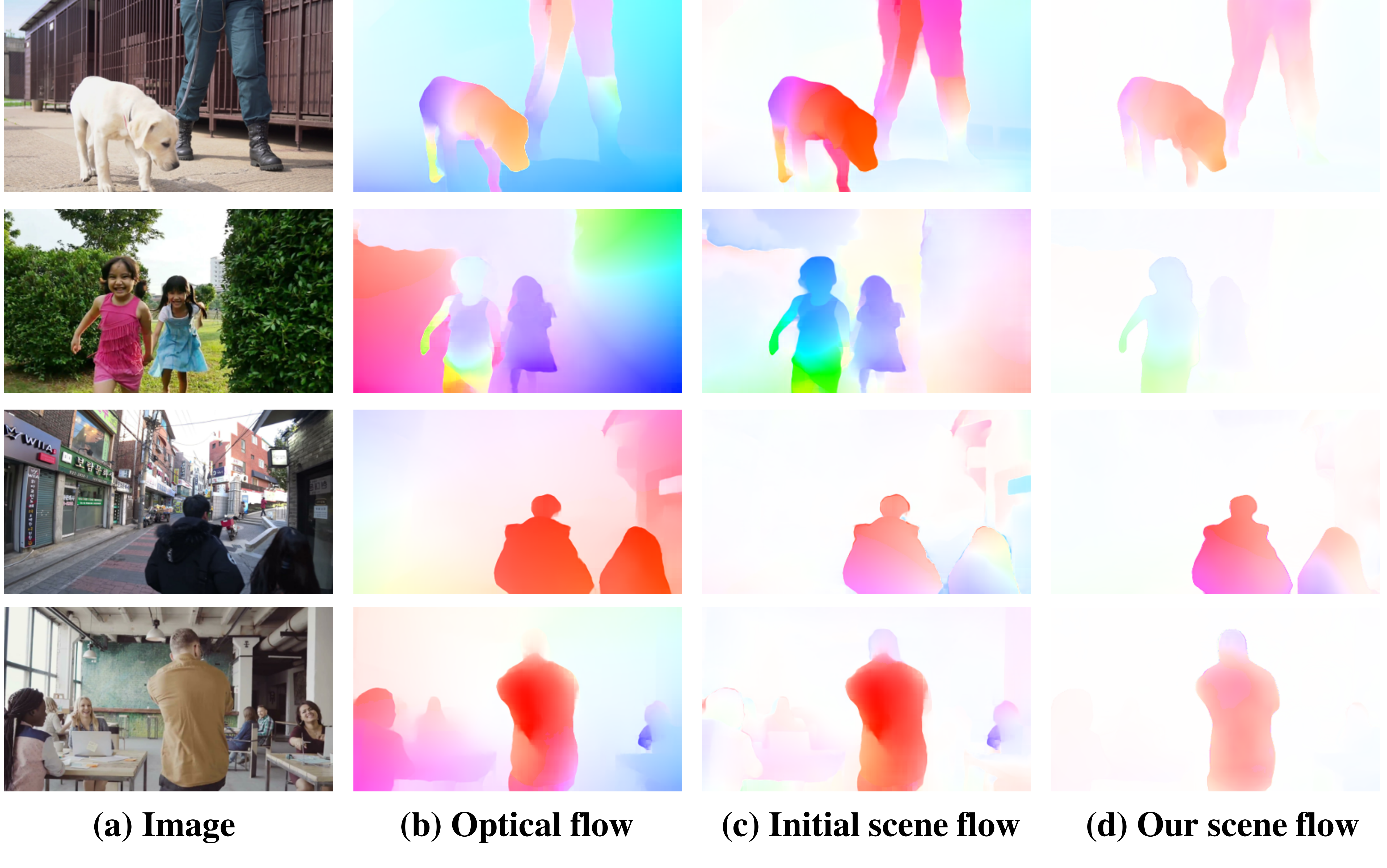}
    \caption{ \textbf{Scene flow results.} Colors correspond to the direction of the projected flow, while saturation corresponds to magnitude (white is zero flow). Optical flow (b) contains both background and foreground motion. Projected scene flow derived from the initial depth estimate (c) and optical flow (b) has small, but still noticeable motion in stationary areas. Our projected scene flow (d) is concentrated in the truly moving regions of the scene.}
    \label{fig:sceneflow_results}
\end{figure}
We test our method on real world videos containing a variety of moving objects (people, pets, cars) performing complex motions such as running or dancing. Representative results are shown in Fig.~\ref{fig:res_real}, and the full set of videos is provided in the SM. All of our results on real videos are produced with MiDaS~\cite{Ranftl2020} initialization.
In addition to depth maps, we also present depth $x\text{-}t$ slices (Fig.~\ref{fig:res_real}f) for qualitative comparison of temporal consistency of our results compared to the initial depth (MiDas) and CVD.

While the per-frame depth maps produced by MiDaS (Fig.~\ref{fig:res_real}c) are generally sharp and  accurate in terms of overall depth ordering, our method shows improved temporal consistency as shown by the reduced zigzagging patterns and color variation in the $x\text{-}t$ slices.

The full-video method CVD produces good temporal consistency and can reconstruct some small motions (e.g. waving hand, Fig.~\ref{fig:res_real} bottom). However, CVD tends to push large moving regions to infinity (Fig.~\ref{fig:res_real}d) due to the violation of epipolar constraints, while our method works well for the moving regions as well as the static regions. Note that no input segmentation of moving vs. static regions is used for these results. The CVD results shown in Fig.~\ref{fig:res_real} is produced by the original implementation by the authors, using the MannequinChallenge model as initialization. CVD with MiDaS initialization fail to train on some of the sequences due to reasons described in Sec.~\ref{sec:sintel_score}

\begin{figure}
     \centering
     \footnotesize
     \begin{tabular}{ccc}
     \includegraphics[width=0.3\columnwidth]{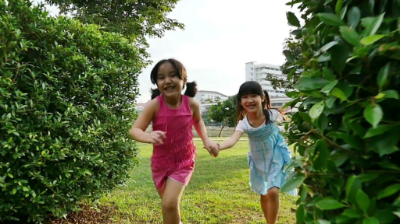} &
     \includegraphics[width=0.3\columnwidth]{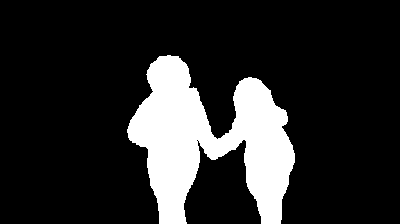} & \\
     Input RGB & Input Motion Mask & \\
     \includegraphics[width=0.3\columnwidth]{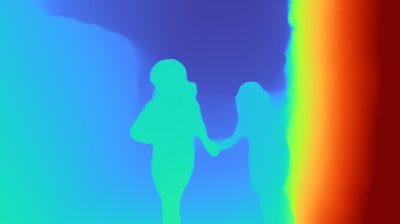} &
     \includegraphics[width=0.3\columnwidth]{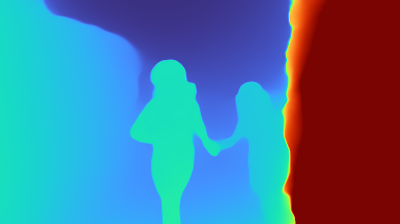} &
     \includegraphics[width=0.3\columnwidth]{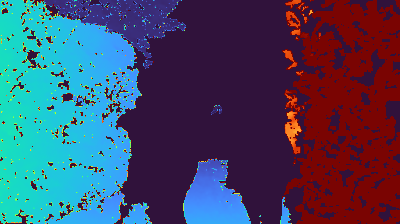}
     \\
     w/o Mask & w/ Mask & COLMAP \\
     \includegraphics[width=0.3\columnwidth]{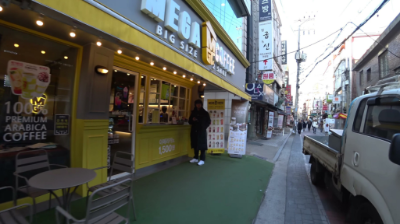} &
     \includegraphics[width=0.3\columnwidth]{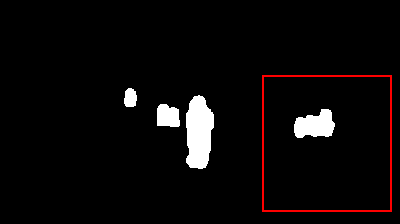} & \\
     Input RGB & Input Motion Mask & \\
     \includegraphics[width=0.3\columnwidth]{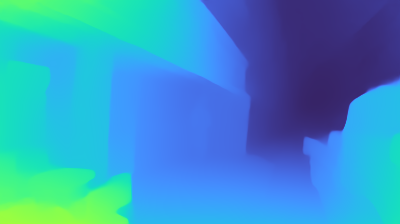} &
     \includegraphics[width=0.3\columnwidth]{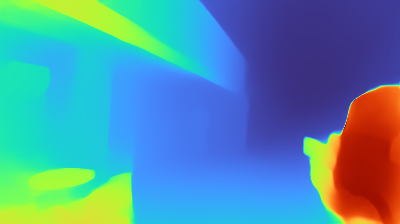} &
     \includegraphics[width=0.3\columnwidth]{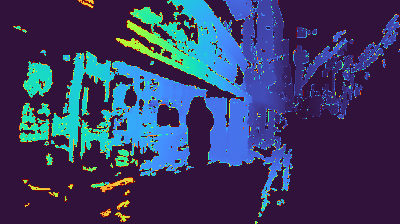}
     \\
     w/o Mask & w/ Mask & COLMAP \\
     \end{tabular}
     \normalsize
     \caption{{\bf Results with a motion mask.} When an accurate motion mask is available as input, our method can incorporate it to improve static regions (top): note the hedge on right and sharper background. However, if the motion mask is inaccurate, it can negatively affect the results (bottom): note truck is not marked as moving (red box) and is spuriously pulled towards the camera. 
     }
     \label{fig:motion_mask}
\end{figure}

The output of the scene flow MLP is visualized in Fig.~\ref{fig:sceneflow_results} by projecting to image space. The scene flow of our method (d) is almost zero on static regions, while scene flows derived from the initial depth estimates (c) assign significant motion to stationary objects, in some cases of similar magnitude as moving objects (ground and wall in first row, hedge and sky in second row). This spurious motion stems from the lack of temporal coherence in the initial depth maps. Note that our method produces accurate, non-zero scene flow even when optical flow is close to zero (head of left person in second row, center person in bottom row). 

\begin{figure*}
     \centering
     \footnotesize
     \begin{tabular}{cccc}
     \includegraphics[width=0.5\columnwidth]{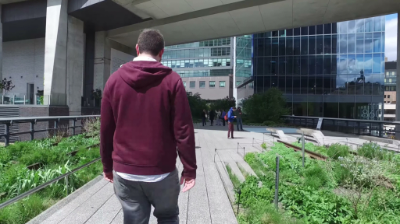} &
     \includegraphics[width=0.5\columnwidth]{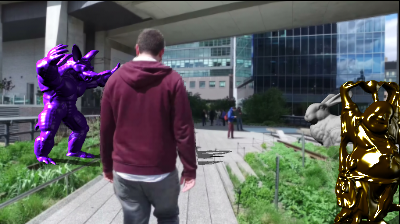} &
     \includegraphics[width=0.5\columnwidth]{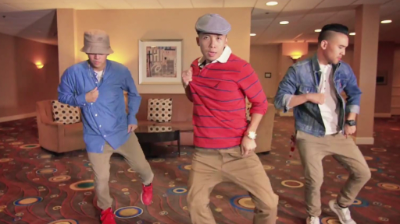} &
     \includegraphics[width=0.5\columnwidth]{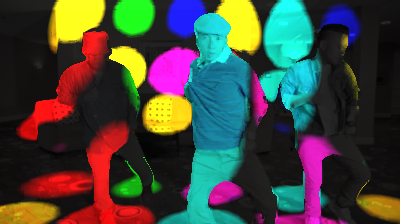} \\
     \includegraphics[width=0.5\columnwidth]{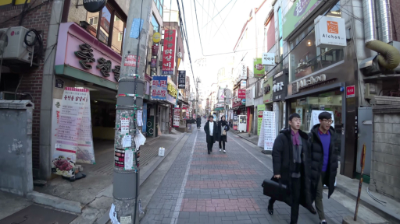} &
     \includegraphics[width=0.5\columnwidth]{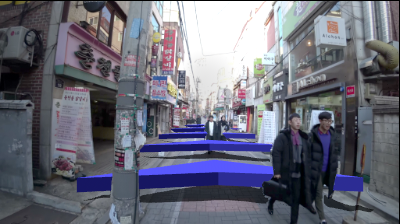} &
     \includegraphics[width=0.5\columnwidth]{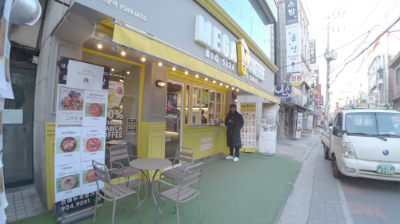} & 
     \includegraphics[width=0.5\columnwidth]{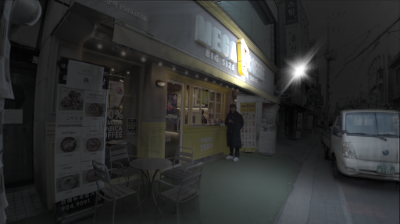}\\
     Original & Insertion Result & Original & Relighting Result \\
     \end{tabular}
     \normalsize
     \vspace{-0.5em}
     \caption{{\bf Object and light insertion.} Because of their temporal coherence, the depth buffers produced by our method may be used to insert objects (top) or position-dependent lighting effects (bottom) without distracting flickering. Please see supplemental material for a video comparison with single-frame depth prediction.
     }
     \vspace{-0.5em}
     \label{fig:insertion_results}
\end{figure*}
When an accurate motion mask is available for a real video, our method can incorporate it to improve results in static parts of the scene (Fig.~\ref{fig:motion_mask}, top). Using an inaccurate motion mask can harm results, however (Fig.~\ref{fig:motion_mask}, bottom): the truck is incorrectly marked as stationary, and is spuriously pulled towards the camera as a result. We do not use motion masks for any results other than Fig~\ref{fig:motion_mask}.

\subsection{Applications}
\label{sec:applications}

We explore several applications of our depth maps, including inserting  virtual objects into a video, inserting a new light source that travels the scene, or segmenting the video via tracking of simple proxy geometries in 3D. In all these applications, the geometrical and temporal consistency of the depth maps play a critical role in producing  realistic effects in which  depth ordering of moving objects is preserved and the result is consistent over time.

\paragraph{Object and Light Insertion}
Fig.~\ref{fig:insertion_results} shows the results of inserting synthetic objects and lights into real-world videos. The 3D scene is created by unprojecting our depth maps per-frame using the per-frame input camera, without further 3D processing. To slightly improve the sharpness of depth edges, we identify edges between moving and static regions using the depth map and projected scene flow, then separately apply morphological erosion followed by dilation to the moving and static regions in order to replace values along depth edges with depth values from interior regions. Lights and objects were placed manually using NUKE~\cite{nuke}.

Temporal coherence is critical for creating effective insertion results: while the human eye is often not sensitive to small inaccuracies in shadow or lighting placement, it is extremely sensitive to the jitter or flashing that can occur when the depth predictions near an inserted object are unstable. The depth maps produced by our method are stable enough to avoid distracting artifacts compared with single-frame depth prediction, even with large moving objects (see supplementary video for comparison).

\paragraph{3D Segmentation}

\begin{figure}
     \centering
     \footnotesize
     \begin{tabular}{ccc}
     \includegraphics[width=0.3\columnwidth]{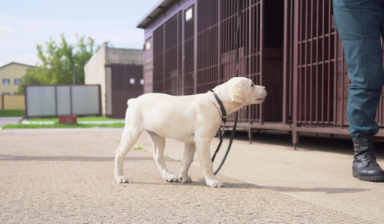} &
     \includegraphics[width=0.3\columnwidth]{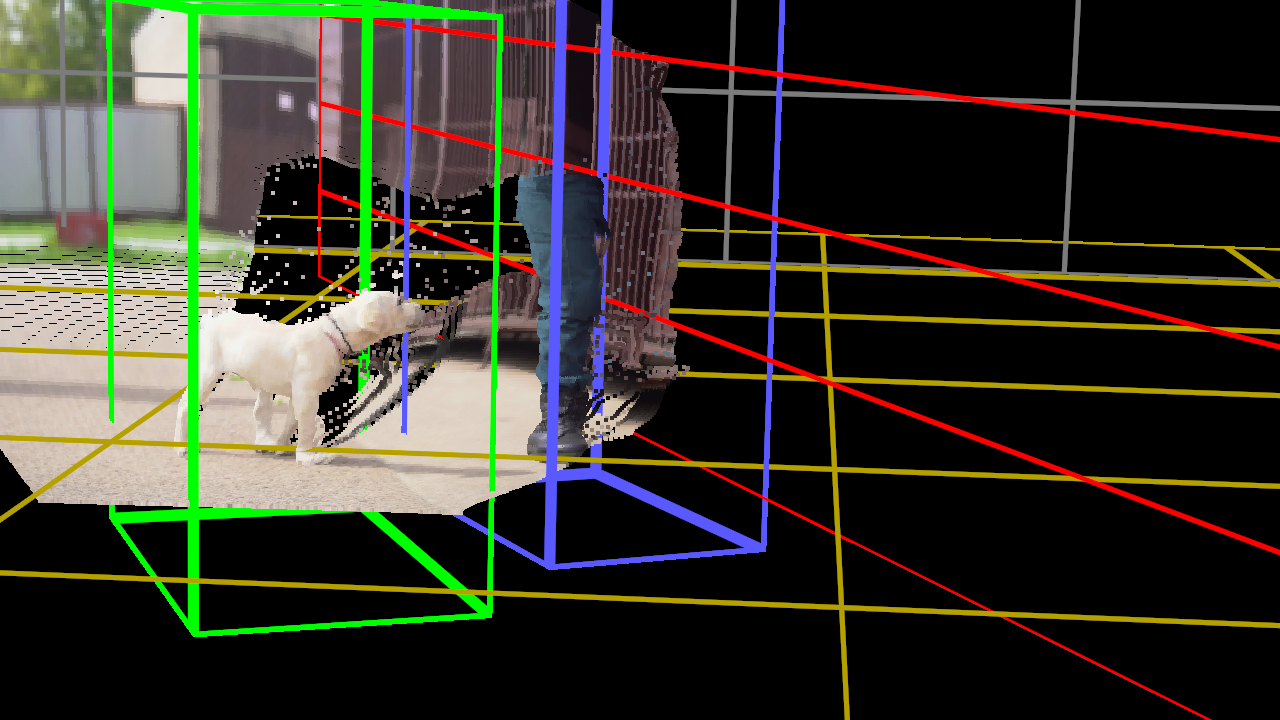} &
     \includegraphics[width=0.3\columnwidth]{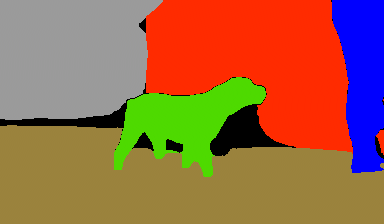} \\
     \includegraphics[width=0.3\columnwidth]{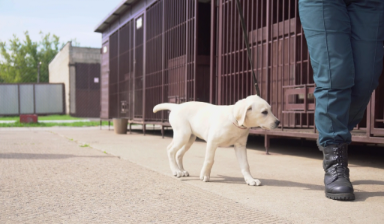} &
     \includegraphics[width=0.3\columnwidth]{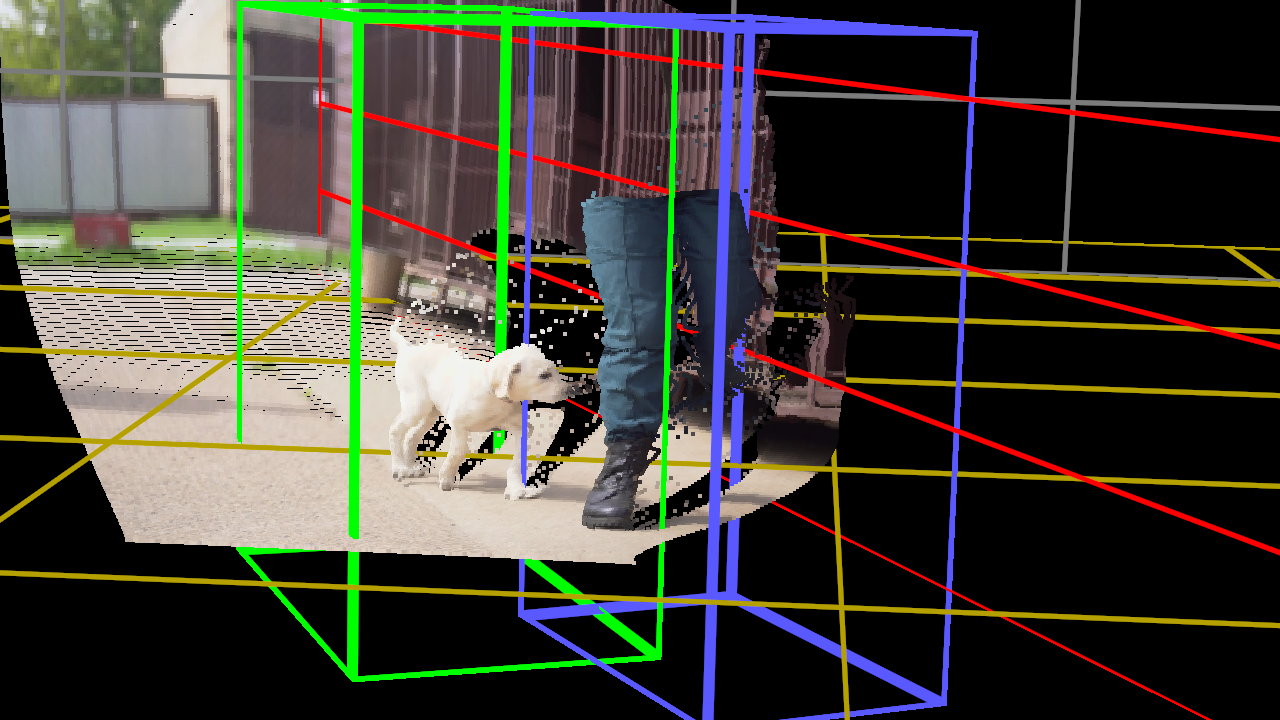} &
     \includegraphics[width=0.3\columnwidth]{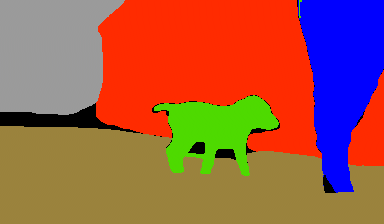} \\
     \includegraphics[width=0.3\columnwidth]{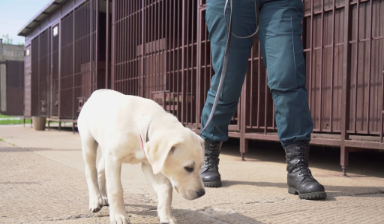} &
     \includegraphics[width=0.3\columnwidth]{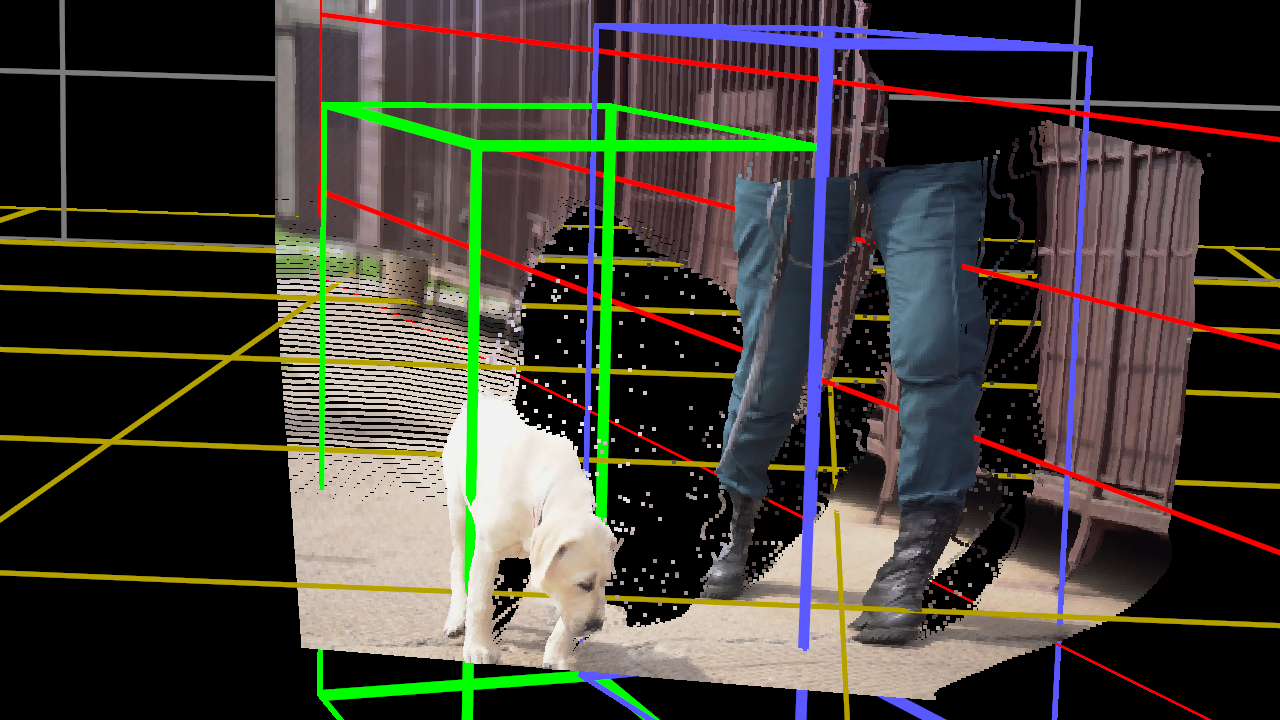} &
     \includegraphics[width=0.3\columnwidth]{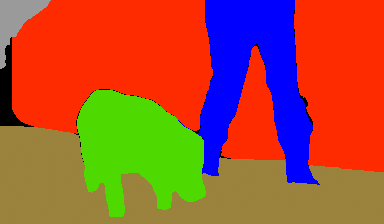} \\
     Input & 3D Proxy Geometry & Segmentation \\
     \end{tabular}
     \normalsize
     \vspace{-0.5em}
     \caption{{\bf Image segmentation using 3D proxy geometry.} An input image (left) may be segmented by using simple proxy geometry in 3D (center). Points that lie within boxes (dog, person) or behind planes (ground, wall, background) are colored by the respective proxies' color to produce the segmentation (right). Black indicates areas outside of a proxy or inside multiple proxies. The green and blue boxes track the dog and person using manual keyframes. Because of the temporal coherence of the depth maps, keyframes were needed on only 8 of 121 frames for this sequence.
     }
     \vspace{-0.5em}
     \label{fig:segmentation_results}
\end{figure}

The stability of our depth maps allows for effective manual segmentation of videos using 3D information. The user may place simple proxy geometry around the objects in 3D, then color the points by the proxy geometry they fall inside (Fig.~\ref{fig:segmentation_results}). The proxies may be animated to track moving objects such as the dog and walking person (Fig.~\ref{fig:segmentation_results}, left). Loose proxy geometry usually suffices to separate the objects in 3D: for the ``puppy'' sequence, each segment was represented by either a box (dog, person) or a plane (ground, wall, background). The position of the bounding boxes was manually adjusted over time. Only 8 keyframes for the 121 frame sequence were required. This 3D setup can be achieved in a few minutes using a 3D video editing package such as NUKE~\cite{nuke}. 

Unlike data-drive segmentation models such as Mask R-CNN~\cite{he2017mask}, this approach allows the user to segment objects that may not correspond to a specific label in the segmentation training set (e.g., the red wall in Fig.~\ref{fig:segmentation_results}, right). To obtain this result using learned, image-based segmentation, a dataset of ground-truth segmentations with labels corresponding to geometric features would typically be required.

\section{Discussion and Limitations}

Our method generates high-quality, temporally consistent depth maps for arbitrary moving objects in video, suitable for effects such as object insertion, relighting, and 3D segmentation. There are two main areas for improvement, however: failures due to inaccuracy of optical flow, and accuracy of occlusion boundaries.

\begin{figure}
     \centering
     \footnotesize
     \begin{tabular}{cccc}
     \includegraphics[width=0.23\columnwidth]{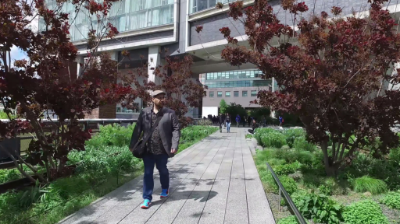} &
     \hspace{-1em}\includegraphics[width=0.23\columnwidth]{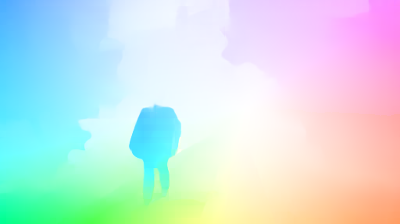} &
     \hspace{-1em}\includegraphics[width=0.23\columnwidth]{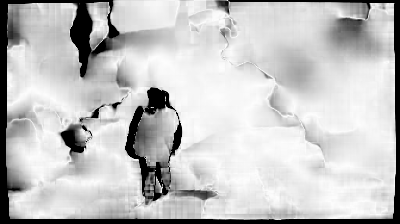} &
     \hspace{-1em}\includegraphics[width=0.23\columnwidth]{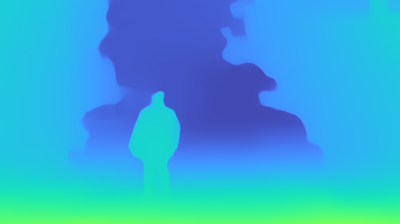} \\
     \includegraphics[width=0.23\columnwidth]{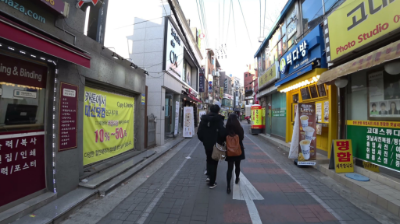} &
     \hspace{-1em}\includegraphics[width=0.23\columnwidth]{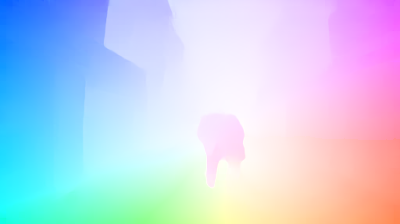} &
     \hspace{-1em}\includegraphics[width=0.23\columnwidth]{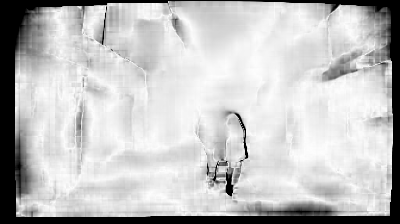} &
     \hspace{-1em}\includegraphics[width=0.23\columnwidth]{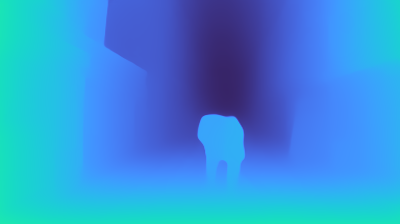} \\
     Input & \hspace{-1em}Optical Flow & \hspace{-1em}Flow Consistency & \hspace{-1em}Our Result \\
     \end{tabular}
     \normalsize
     \caption{{\bf Effect of optical flow failure.} Even when evaluated with wide baselines, RAFT optical flow~\cite{teed2020raft} may be inaccurate. Where our mask $\occlusion{i}{i+k}$ marks flow as inaccurate, our depth estimates are robust to flow errors (top: note head of person is dark in consistency map). If, however, flow is self-consistent but wrong, our method produces incorrect results (bottom, note heads are marked consistent).
     }
     \label{fig:flow_failures}
\end{figure}

Our performance is affected by the accuracy of the optical flow estimates. While state-of-the art optical flow methods are getting better and better, they still have errors in some cases. Our method can handle  optical flow errors to some extent as long as the erroneous regions can be identified using a left-right consistency check and removed from our objective (as described in Sec.~\ref{sec:losses}). We observed that in some cases,  severe errors such as removal of heads and limbs can still be left-right consistent  (with less than a pixel error). In such cases, the optical flow errors translate into errors in our depth prediction (Fig.~\ref{fig:flow_failures}). This problem can be potentially alleviated  by developing more intelligent confidence measurements, or by exploring the use of photometric reconstruction losses. 

The occlusion boundaries produced by our method may be inaccurate by a few pixels, and so are not yet suitable for demanding effects such as novel view synthesis. Blurring of depth boundaries is a common side effect of CNN depth prediction, and semi-transparent effects such as hair are not solvable using a single depth map. Edge-aware filtering of depth maps~\cite{barron2016fast} may improve sharpness, but may reduce temporal consistency. In future work, we hope to integrate a matting or layering approach (e.g.~\cite{lu2020retiming}) to better handle difficult depth boundaries.

\bibliographystyle{ACM-Reference-Format}
\bibliography{refs.bib}


\begin{thebibliography}{54}


\ifx \showCODEN    \undefined \def \showCODEN     #1{\unskip}     \fi
\ifx \showDOI      \undefined \def \showDOI       #1{#1}\fi
\ifx \showISBNx    \undefined \def \showISBNx     #1{\unskip}     \fi
\ifx \showISBNxiii \undefined \def \showISBNxiii  #1{\unskip}     \fi
\ifx \showISSN     \undefined \def \showISSN      #1{\unskip}     \fi
\ifx \showLCCN     \undefined \def \showLCCN      #1{\unskip}     \fi
\ifx \shownote     \undefined \def \shownote      #1{#1}          \fi
\ifx \showarticletitle \undefined \def \showarticletitle #1{#1}   \fi
\ifx \showURL      \undefined \def \showURL       {\relax}        \fi
\providecommand\bibfield[2]{#2}
\providecommand\bibinfo[2]{#2}
\providecommand\natexlab[1]{#1}
\providecommand\showeprint[2][]{arXiv:#2}

\bibitem[\protect\citeauthoryear{Bansal, Vo, Sheikh, Ramanan, and
  Narasimhan}{Bansal et~al\mbox{.}}{2020}]%
        {bansal20204d}
\bibfield{author}{\bibinfo{person}{Aayush Bansal}, \bibinfo{person}{Minh Vo},
  \bibinfo{person}{Yaser Sheikh}, \bibinfo{person}{Deva Ramanan}, {and}
  \bibinfo{person}{Srinivasa Narasimhan}.} \bibinfo{year}{2020}\natexlab{}.
\newblock \showarticletitle{4d visualization of dynamic events from
  unconstrained multi-view videos}. In \bibinfo{booktitle}{\emph{Proceedings of
  the IEEE/CVF Conference on Computer Vision and Pattern Recognition}}.
  \bibinfo{pages}{5366--5375}.
\newblock


\bibitem[\protect\citeauthoryear{Barron and Poole}{Barron and Poole}{2016}]%
        {barron2016fast}
\bibfield{author}{\bibinfo{person}{Jonathan~T Barron} {and}
  \bibinfo{person}{Ben Poole}.} \bibinfo{year}{2016}\natexlab{}.
\newblock \showarticletitle{The fast bilateral solver}. In
  \bibinfo{booktitle}{\emph{European Conference on Computer Vision}}. Springer,
  \bibinfo{pages}{617--632}.
\newblock


\bibitem[\protect\citeauthoryear{Basha, Avidan, Hornung, and Matusik}{Basha
  et~al\mbox{.}}{2012}]%
        {basha2012structure}
\bibfield{author}{\bibinfo{person}{Tali Basha}, \bibinfo{person}{Shai Avidan},
  \bibinfo{person}{Alexander Hornung}, {and} \bibinfo{person}{Wojciech
  Matusik}.} \bibinfo{year}{2012}\natexlab{}.
\newblock \showarticletitle{Structure and motion from scene registration}. In
  \bibinfo{booktitle}{\emph{IEEE Conf. Comput. Vis. Pattern Recog.}} IEEE.
\newblock


\bibitem[\protect\citeauthoryear{Basha, Moses, and Kiryati}{Basha
  et~al\mbox{.}}{2013}]%
        {basha2013multi}
\bibfield{author}{\bibinfo{person}{Tali Basha}, \bibinfo{person}{Yael Moses},
  {and} \bibinfo{person}{Nahum Kiryati}.} \bibinfo{year}{2013}\natexlab{}.
\newblock \showarticletitle{Multi-view scene flow estimation: A view centered
  variational approach}.
\newblock \bibinfo{journal}{\emph{Int. J. Comput. Vis.}} \bibinfo{number}{1}
  (\bibinfo{year}{2013}).
\newblock


\bibitem[\protect\citeauthoryear{Butler, Wulff, Stanley, and Black}{Butler
  et~al\mbox{.}}{2012}]%
        {sintel}
\bibfield{author}{\bibinfo{person}{D.~J. Butler}, \bibinfo{person}{J. Wulff},
  \bibinfo{person}{G.~B. Stanley}, {and} \bibinfo{person}{M.~J. Black}.}
  \bibinfo{year}{2012}\natexlab{}.
\newblock \showarticletitle{A naturalistic open source movie for optical flow
  evaluation}. In \bibinfo{booktitle}{\emph{European Conf. on Computer Vision
  (ECCV)}} \emph{(\bibinfo{series}{Part IV, LNCS 7577})},
  \bibfield{editor}{\bibinfo{person}{{A. Fitzgibbon et al. (Eds.)}}} (Ed.).
  \bibinfo{publisher}{Springer-Verlag}, \bibinfo{pages}{611--625}.
\newblock


\bibitem[\protect\citeauthoryear{Casser, Pirk, Mahjourian, and Angelova}{Casser
  et~al\mbox{.}}{2019a}]%
        {casser2019unsupervised}
\bibfield{author}{\bibinfo{person}{Vincent Casser}, \bibinfo{person}{Soeren
  Pirk}, \bibinfo{person}{Reza Mahjourian}, {and} \bibinfo{person}{Anelia
  Angelova}.} \bibinfo{year}{2019}\natexlab{a}.
\newblock \showarticletitle{Unsupervised learning of depth and ego-motion: A
  structured approach}. In \bibinfo{booktitle}{\emph{Thirty-Third AAAI
  Conference on Artificial Intelligence (AAAI-19)}}, Vol.~\bibinfo{volume}{2}.
  \bibinfo{pages}{7}.
\newblock


\bibitem[\protect\citeauthoryear{Casser, Pirk, Mahjourian, and Angelova}{Casser
  et~al\mbox{.}}{2019b}]%
        {casser2019depth}
\bibfield{author}{\bibinfo{person}{Vincent~Michael Casser},
  \bibinfo{person}{Soeren Pirk}, \bibinfo{person}{Reza Mahjourian}, {and}
  \bibinfo{person}{Anelia Angelova}.} \bibinfo{year}{2019}\natexlab{b}.
\newblock \showarticletitle{Depth Prediction Without the Sensors: Leveraging
  Structure for Unsupervised Learning from Monocular Videos}. In
  \bibinfo{booktitle}{\emph{AAAI}}.
\newblock


\bibitem[\protect\citeauthoryear{Chen, Fu, Yang, and Deng}{Chen
  et~al\mbox{.}}{2016}]%
        {chen2016single}
\bibfield{author}{\bibinfo{person}{Weifeng Chen}, \bibinfo{person}{Zhao Fu},
  \bibinfo{person}{Dawei Yang}, {and} \bibinfo{person}{Jia Deng}.}
  \bibinfo{year}{2016}\natexlab{}.
\newblock \showarticletitle{Single-image depth perception in the wild}.
\newblock \bibinfo{journal}{\emph{arXiv preprint arXiv:1604.03901}}
  (\bibinfo{year}{2016}).
\newblock


\bibitem[\protect\citeauthoryear{Chen, Schmid, and Sminchisescu}{Chen
  et~al\mbox{.}}{2019}]%
        {chen2019self}
\bibfield{author}{\bibinfo{person}{Yuhua Chen}, \bibinfo{person}{Cordelia
  Schmid}, {and} \bibinfo{person}{Cristian Sminchisescu}.}
  \bibinfo{year}{2019}\natexlab{}.
\newblock \showarticletitle{Self-supervised learning with geometric constraints
  in monocular video: Connecting flow, depth, and camera}. In
  \bibinfo{booktitle}{\emph{Int. Conf. Comput. Vis.}}
  \bibinfo{pages}{7063--7072}.
\newblock


\bibitem[\protect\citeauthoryear{Dou, Khamis, Degtyarev, Davidson, Fanello,
  Kowdle, Orts, Rhemann, Kim, Taylor, Kohli, Tankovich, and Izadi}{Dou
  et~al\mbox{.}}{2016}]%
        {Dou2016Fusion4DRP}
\bibfield{author}{\bibinfo{person}{Mingsong Dou}, \bibinfo{person}{Sameh
  Khamis}, \bibinfo{person}{Yury Degtyarev}, \bibinfo{person}{Philip~L.
  Davidson}, \bibinfo{person}{Sean~Ryan Fanello}, \bibinfo{person}{Adarsh
  Kowdle}, \bibinfo{person}{Sergio Orts}, \bibinfo{person}{Christoph Rhemann},
  \bibinfo{person}{David Kim}, \bibinfo{person}{Jonathan Taylor},
  \bibinfo{person}{Pushmeet Kohli}, \bibinfo{person}{Vladimir Tankovich}, {and}
  \bibinfo{person}{Shahram Izadi}.} \bibinfo{year}{2016}\natexlab{}.
\newblock \showarticletitle{{Fusion4D}: real-time performance capture of
  challenging scenes}.
\newblock \bibinfo{journal}{\emph{ACM Trans. Graph.}}  \bibinfo{volume}{35}
  (\bibinfo{year}{2016}).
\newblock


\bibitem[\protect\citeauthoryear{Eigen, Puhrsch, and Fergus}{Eigen
  et~al\mbox{.}}{2014}]%
        {eigen2014depth}
\bibfield{author}{\bibinfo{person}{David Eigen}, \bibinfo{person}{Christian
  Puhrsch}, {and} \bibinfo{person}{Rob Fergus}.}
  \bibinfo{year}{2014}\natexlab{}.
\newblock \showarticletitle{Depth map prediction from a single image using a
  multi-scale deep network}.
\newblock \bibinfo{journal}{\emph{Neural Information Processing Systems}}
  (\bibinfo{year}{2014}).
\newblock


\bibitem[\protect\citeauthoryear{Fu, Gong, Wang, Batmanghelich, and Tao}{Fu
  et~al\mbox{.}}{2018}]%
        {fu2018deep}
\bibfield{author}{\bibinfo{person}{Huan Fu}, \bibinfo{person}{Mingming Gong},
  \bibinfo{person}{Chaohui Wang}, \bibinfo{person}{Kayhan Batmanghelich}, {and}
  \bibinfo{person}{Dacheng Tao}.} \bibinfo{year}{2018}\natexlab{}.
\newblock \showarticletitle{Deep ordinal regression network for monocular depth
  estimation}. In \bibinfo{booktitle}{\emph{Proceedings of the IEEE Conference
  on Computer Vision and Pattern Recognition}}. \bibinfo{pages}{2002--2011}.
\newblock


\bibitem[\protect\citeauthoryear{Godard, Mac~Aodha, and Brostow}{Godard
  et~al\mbox{.}}{2017}]%
        {godard2017unsupervised}
\bibfield{author}{\bibinfo{person}{Cl{\'e}ment Godard}, \bibinfo{person}{Oisin
  Mac~Aodha}, {and} \bibinfo{person}{Gabriel~J Brostow}.}
  \bibinfo{year}{2017}\natexlab{}.
\newblock \showarticletitle{Unsupervised monocular depth estimation with
  left-right consistency}. In \bibinfo{booktitle}{\emph{Proceedings of the IEEE
  Conference on Computer Vision and Pattern Recognition}}.
  \bibinfo{pages}{270--279}.
\newblock


\bibitem[\protect\citeauthoryear{Godard, {Mac Aodha}, Firman, and
  Brostow}{Godard et~al\mbox{.}}{2019}]%
        {monodepth2}
\bibfield{author}{\bibinfo{person}{Cl{\'{e}}ment Godard},
  \bibinfo{person}{Oisin {Mac Aodha}}, \bibinfo{person}{Michael Firman}, {and}
  \bibinfo{person}{Gabriel~J. Brostow}.} \bibinfo{year}{2019}\natexlab{}.
\newblock \showarticletitle{Digging into Self-Supervised Monocular Depth
  Prediction}.
\newblock  (\bibinfo{date}{October} \bibinfo{year}{2019}).
\newblock


\bibitem[\protect\citeauthoryear{He, Gkioxari, Doll{\'a}r, and Girshick}{He
  et~al\mbox{.}}{2017}]%
        {he2017mask}
\bibfield{author}{\bibinfo{person}{Kaiming He}, \bibinfo{person}{Georgia
  Gkioxari}, \bibinfo{person}{Piotr Doll{\'a}r}, {and} \bibinfo{person}{Ross
  Girshick}.} \bibinfo{year}{2017}\natexlab{}.
\newblock \showarticletitle{Mask r-cnn}. In
  \bibinfo{booktitle}{\emph{Proceedings of the IEEE international conference on
  computer vision}}. \bibinfo{pages}{2961--2969}.
\newblock


\bibitem[\protect\citeauthoryear{Innmann, Zollh{\"o}fer, Niessner, Theobalt,
  and Stamminger}{Innmann et~al\mbox{.}}{2016}]%
        {Innmann2016VolumeDeformRV}
\bibfield{author}{\bibinfo{person}{Matthias Innmann}, \bibinfo{person}{Michael
  Zollh{\"o}fer}, \bibinfo{person}{Matthias Niessner},
  \bibinfo{person}{Christian Theobalt}, {and} \bibinfo{person}{Marc
  Stamminger}.} \bibinfo{year}{2016}\natexlab{}.
\newblock \showarticletitle{{VolumeDeform}: {R}eal-time Volumetric Non-rigid
  Reconstruction}. In \bibinfo{booktitle}{\emph{Eur. Conf. Comput. Vis.}}
\newblock


\bibitem[\protect\citeauthoryear{Jensen, Doest, Aan{\ae}s, and Del~Bue}{Jensen
  et~al\mbox{.}}{2020}]%
        {jensen2020benchmark}
\bibfield{author}{\bibinfo{person}{Sebastian Hoppe~Nesgaard Jensen},
  \bibinfo{person}{Mads Emil~Brix Doest}, \bibinfo{person}{Henrik Aan{\ae}s},
  {and} \bibinfo{person}{Alessio Del~Bue}.} \bibinfo{year}{2020}\natexlab{}.
\newblock \showarticletitle{A benchmark and evaluation of non-rigid structure
  from motion}.
\newblock \bibinfo{journal}{\emph{International Journal of Computer Vision}}
  (\bibinfo{year}{2020}), \bibinfo{pages}{1--18}.
\newblock


\bibitem[\protect\citeauthoryear{Klingner, Term{\"o}hlen, Mikolajczyk, and
  Fingscheidt}{Klingner et~al\mbox{.}}{2020}]%
        {klingner2020self}
\bibfield{author}{\bibinfo{person}{Marvin Klingner}, \bibinfo{person}{Jan-Aike
  Term{\"o}hlen}, \bibinfo{person}{Jonas Mikolajczyk}, {and}
  \bibinfo{person}{Tim Fingscheidt}.} \bibinfo{year}{2020}\natexlab{}.
\newblock \showarticletitle{Self-supervised monocular depth estimation: Solving
  the dynamic object problem by semantic guidance}. In
  \bibinfo{booktitle}{\emph{European Conference on Computer Vision}}.
  \bibinfo{pages}{582--600}.
\newblock


\bibitem[\protect\citeauthoryear{Kopf, Rong, and Huang}{Kopf
  et~al\mbox{.}}{2020}]%
        {kopf2020robust}
\bibfield{author}{\bibinfo{person}{Johannes Kopf}, \bibinfo{person}{Xuejian
  Rong}, {and} \bibinfo{person}{Jia-Bin Huang}.}
  \bibinfo{year}{2020}\natexlab{}.
\newblock \showarticletitle{Robust Consistent Video Depth Estimation}.
\newblock \bibinfo{journal}{\emph{arXiv preprint arXiv:2012.05901}}
  (\bibinfo{year}{2020}).
\newblock


\bibitem[\protect\citeauthoryear{Li, Dekel, Cole, Tucker, Snavely, Liu, and
  Freeman}{Li et~al\mbox{.}}{2019}]%
        {li2019learning}
\bibfield{author}{\bibinfo{person}{Zhengqi Li}, \bibinfo{person}{Tali Dekel},
  \bibinfo{person}{Forrester Cole}, \bibinfo{person}{Richard Tucker},
  \bibinfo{person}{Noah Snavely}, \bibinfo{person}{Ce Liu}, {and}
  \bibinfo{person}{William~T Freeman}.} \bibinfo{year}{2019}\natexlab{}.
\newblock \showarticletitle{Learning the depths of moving people by watching
  frozen people}. In \bibinfo{booktitle}{\emph{IEEE Conf. Comput. Vis. Pattern
  Recog.}}
\newblock


\bibitem[\protect\citeauthoryear{Li, Dekel, Cole, Tucker, Snavely, Liu, and
  Freeman}{Li et~al\mbox{.}}{2020a}]%
        {li2020mannequinchallenge}
\bibfield{author}{\bibinfo{person}{Zhengqi Li}, \bibinfo{person}{Tali Dekel},
  \bibinfo{person}{Forrester Cole}, \bibinfo{person}{Richard Tucker},
  \bibinfo{person}{Noah Snavely}, \bibinfo{person}{Ce Liu}, {and}
  \bibinfo{person}{William~T Freeman}.} \bibinfo{year}{2020}\natexlab{a}.
\newblock \showarticletitle{MannequinChallenge: Learning the Depths of Moving
  People by Watching Frozen People}.
\newblock \bibinfo{journal}{\emph{IEEE Trans. Pattern Anal. Mach. Intell.}}
  (\bibinfo{year}{2020}).
\newblock


\bibitem[\protect\citeauthoryear{Li, Niklaus, Snavely, and Wang}{Li
  et~al\mbox{.}}{2020b}]%
        {li2020neural}
\bibfield{author}{\bibinfo{person}{Zhengqi Li}, \bibinfo{person}{Simon
  Niklaus}, \bibinfo{person}{Noah Snavely}, {and} \bibinfo{person}{Oliver
  Wang}.} \bibinfo{year}{2020}\natexlab{b}.
\newblock \showarticletitle{Neural Scene Flow Fields for Space-Time View
  Synthesis of Dynamic Scenes}.
\newblock \bibinfo{journal}{\emph{arXiv preprint arXiv:2011.13084}}
  (\bibinfo{year}{2020}).
\newblock


\bibitem[\protect\citeauthoryear{Li and Snavely}{Li and Snavely}{2018}]%
        {li2018megadepth}
\bibfield{author}{\bibinfo{person}{Zhengqi Li} {and} \bibinfo{person}{Noah
  Snavely}.} \bibinfo{year}{2018}\natexlab{}.
\newblock \showarticletitle{Megadepth: Learning single-view depth prediction
  from internet photos}. In \bibinfo{booktitle}{\emph{Proceedings of the IEEE
  Conference on Computer Vision and Pattern Recognition}}.
  \bibinfo{pages}{2041--2050}.
\newblock


\bibitem[\protect\citeauthoryear{Ltd}{Ltd}{2018}]%
        {nuke}
\bibfield{author}{\bibinfo{person}{The Foundry~Visionmongers Ltd}.}
  \bibinfo{year}{2018}\natexlab{}.
\newblock \bibinfo{booktitle}{\emph{NUKE}}.
\newblock
\urldef\tempurl%
\url{https://www.foundry.com/products/nuke}
\showURL{%
\tempurl}


\bibitem[\protect\citeauthoryear{Lu, Cole, Dekel, Xie, Zisserman, Salesin,
  Freeman, and Rubinstein}{Lu et~al\mbox{.}}{2020}]%
        {lu2020retiming}
\bibfield{author}{\bibinfo{person}{Erika Lu}, \bibinfo{person}{Forrester Cole},
  \bibinfo{person}{Tali Dekel}, \bibinfo{person}{Weidi Xie},
  \bibinfo{person}{Andrew Zisserman}, \bibinfo{person}{David Salesin},
  \bibinfo{person}{William~T. Freeman}, {and} \bibinfo{person}{Michael
  Rubinstein}.} \bibinfo{year}{2020}\natexlab{}.
\newblock \showarticletitle{Layered Neural Rendering for Retiming People in
  Video}.
\newblock \bibinfo{journal}{\emph{ACM Trans. Graph.}} \bibinfo{volume}{39},
  \bibinfo{number}{6}, Article \bibinfo{articleno}{256} (\bibinfo{date}{Nov.}
  \bibinfo{year}{2020}), \bibinfo{numpages}{14}~pages.
\newblock


\bibitem[\protect\citeauthoryear{Luo, Huang, Szeliski, Matzen, and Kopf}{Luo
  et~al\mbox{.}}{2020}]%
        {Luo-VideoDepth-2020}
\bibfield{author}{\bibinfo{person}{Xuan Luo}, \bibinfo{person}{Jia{-}Bin
  Huang}, \bibinfo{person}{Richard Szeliski}, \bibinfo{person}{Kevin Matzen},
  {and} \bibinfo{person}{Johannes Kopf}.} \bibinfo{year}{2020}\natexlab{}.
\newblock \showarticletitle{Consistent Video Depth Estimation}.
\newblock  (\bibinfo{year}{2020}).
\newblock


\bibitem[\protect\citeauthoryear{Mildenhall, Srinivasan, Tancik, Barron,
  Ramamoorthi, and Ng}{Mildenhall et~al\mbox{.}}{2020}]%
        {mildenhall2020nerf}
\bibfield{author}{\bibinfo{person}{Ben Mildenhall}, \bibinfo{person}{Pratul~P.
  Srinivasan}, \bibinfo{person}{Matthew Tancik}, \bibinfo{person}{Jonathan~T.
  Barron}, \bibinfo{person}{Ravi Ramamoorthi}, {and} \bibinfo{person}{Ren Ng}.}
  \bibinfo{year}{2020}\natexlab{}.
\newblock \showarticletitle{NeRF: Representing Scenes as Neural Radiance Fields
  for View Synthesis}. In \bibinfo{booktitle}{\emph{ECCV}}.
\newblock


\bibitem[\protect\citeauthoryear{Mur-Artal, Montiel, and Tard\'os}{Mur-Artal
  et~al\mbox{.}}{2015}]%
        {murTRO2015}
\bibfield{author}{\bibinfo{person}{Ra\'ul Mur-Artal}, \bibinfo{person}{J.~M.~M.
  Montiel}, {and} \bibinfo{person}{Juan~D. Tard\'os}.}
  \bibinfo{year}{2015}\natexlab{}.
\newblock \showarticletitle{{ORB-SLAM}: a Versatile and Accurate Monocular
  {SLAM} System}.
\newblock \bibinfo{journal}{\emph{IEEE Transactions on Robotics}}
  \bibinfo{volume}{31}, \bibinfo{number}{5} (\bibinfo{year}{2015}),
  \bibinfo{pages}{1147--1163}.
\newblock
\urldef\tempurl%
\url{https://doi.org/10.1109/TRO.2015.2463671}
\showDOI{\tempurl}


\bibitem[\protect\citeauthoryear{Newcombe, Fox, and Seitz}{Newcombe
  et~al\mbox{.}}{2015}]%
        {newcombe2015dynamicfusion}
\bibfield{author}{\bibinfo{person}{Richard~A Newcombe}, \bibinfo{person}{Dieter
  Fox}, {and} \bibinfo{person}{Steven~M Seitz}.}
  \bibinfo{year}{2015}\natexlab{}.
\newblock \showarticletitle{{DynamicFusion: Reconstruction and tracking of
  non-rigid scenes in real-time}}. In \bibinfo{booktitle}{\emph{IEEE Conf.
  Comput. Vis. Pattern Recog.}}
\newblock


\bibitem[\protect\citeauthoryear{Niemeyer, Mescheder, Oechsle, and
  Geiger}{Niemeyer et~al\mbox{.}}{2019}]%
        {Niemeyer2019ICCV}
\bibfield{author}{\bibinfo{person}{Michael Niemeyer}, \bibinfo{person}{Lars
  Mescheder}, \bibinfo{person}{Michael Oechsle}, {and} \bibinfo{person}{Andreas
  Geiger}.} \bibinfo{year}{2019}\natexlab{}.
\newblock \showarticletitle{Occupancy Flow: 4D Reconstruction by Learning
  Particle Dynamics}. In \bibinfo{booktitle}{\emph{International Conference on
  Computer Vision (ICCV)}}.
\newblock


\bibitem[\protect\citeauthoryear{Park, Shiratori, Matthews, and Sheikh}{Park
  et~al\mbox{.}}{2010a}]%
        {park20103d}
\bibfield{author}{\bibinfo{person}{Hyun~Soo Park}, \bibinfo{person}{Takaaki
  Shiratori}, \bibinfo{person}{Iain Matthews}, {and} \bibinfo{person}{Yaser
  Sheikh}.} \bibinfo{year}{2010}\natexlab{a}.
\newblock \showarticletitle{3D reconstruction of a moving point from a series
  of 2D projections}. In \bibinfo{booktitle}{\emph{European conference on
  computer vision}}. Springer, \bibinfo{pages}{158--171}.
\newblock


\bibitem[\protect\citeauthoryear{Park, Shiratori, Matthews, and Sheikh}{Park
  et~al\mbox{.}}{2010b}]%
        {Park20103DRO}
\bibfield{author}{\bibinfo{person}{Hyun~Soo Park}, \bibinfo{person}{Takaaki
  Shiratori}, \bibinfo{person}{Iain~A. Matthews}, {and} \bibinfo{person}{Yaser
  Sheikh}.} \bibinfo{year}{2010}\natexlab{b}.
\newblock \showarticletitle{{3D Reconstruction of a Moving Point from a Series
  of 2D Projections}}. In \bibinfo{booktitle}{\emph{Eur. Conf. Comput. Vis.}}
\newblock


\bibitem[\protect\citeauthoryear{Park, Sinha, Barron, Bouaziz, Goldman, Seitz,
  and Martin-Brualla}{Park et~al\mbox{.}}{2020}]%
        {park2020nerfies}
\bibfield{author}{\bibinfo{person}{Keunhong Park}, \bibinfo{person}{Utkarsh
  Sinha}, \bibinfo{person}{Jonathan~T. Barron}, \bibinfo{person}{Sofien
  Bouaziz}, \bibinfo{person}{Dan~B Goldman}, \bibinfo{person}{Steven~M. Seitz},
  {and} \bibinfo{person}{Ricardo Martin-Brualla}.}
  \bibinfo{year}{2020}\natexlab{}.
\newblock \showarticletitle{Deformable Neural Radiance Fields}.
\newblock \bibinfo{journal}{\emph{arXiv preprint arXiv:2011.12948}}
  (\bibinfo{year}{2020}).
\newblock


\bibitem[\protect\citeauthoryear{Patil, Van~Gansbeke, Dai, and Van~Gool}{Patil
  et~al\mbox{.}}{2020}]%
        {patil2020don}
\bibfield{author}{\bibinfo{person}{Vaishakh Patil}, \bibinfo{person}{Wouter
  Van~Gansbeke}, \bibinfo{person}{Dengxin Dai}, {and} \bibinfo{person}{Luc
  Van~Gool}.} \bibinfo{year}{2020}\natexlab{}.
\newblock \showarticletitle{Don’t forget the past: Recurrent depth estimation
  from monocular video}.
\newblock \bibinfo{journal}{\emph{IEEE Robotics and Automation Letters}}
  \bibinfo{volume}{5}, \bibinfo{number}{4} (\bibinfo{year}{2020}),
  \bibinfo{pages}{6813--6820}.
\newblock


\bibitem[\protect\citeauthoryear{Ranftl, Lasinger, Hafner, Schindler, and
  Koltun}{Ranftl et~al\mbox{.}}{2020}]%
        {Ranftl2020}
\bibfield{author}{\bibinfo{person}{Ren\'{e} Ranftl}, \bibinfo{person}{Katrin
  Lasinger}, \bibinfo{person}{David Hafner}, \bibinfo{person}{Konrad
  Schindler}, {and} \bibinfo{person}{Vladlen Koltun}.}
  \bibinfo{year}{2020}\natexlab{}.
\newblock \showarticletitle{Towards Robust Monocular Depth Estimation: Mixing
  Datasets for Zero-shot Cross-dataset Transfer}.
\newblock \bibinfo{journal}{\emph{IEEE Transactions on Pattern Analysis and
  Machine Intelligence (TPAMI)}} (\bibinfo{year}{2020}).
\newblock


\bibitem[\protect\citeauthoryear{Ranftl, Vineet, Chen, and Koltun}{Ranftl
  et~al\mbox{.}}{2016}]%
        {ranftl2016dense}
\bibfield{author}{\bibinfo{person}{Rene Ranftl}, \bibinfo{person}{Vibhav
  Vineet}, \bibinfo{person}{Qifeng Chen}, {and} \bibinfo{person}{Vladlen
  Koltun}.} \bibinfo{year}{2016}\natexlab{}.
\newblock \showarticletitle{Dense monocular depth estimation in complex dynamic
  scenes}. In \bibinfo{booktitle}{\emph{IEEE Conf. Comput. Vis. Pattern
  Recog.}}
\newblock


\bibitem[\protect\citeauthoryear{Rematas, Kemelmacher-Shlizerman, Curless, and
  Seitz}{Rematas et~al\mbox{.}}{2018}]%
        {rematas2018soccer}
\bibfield{author}{\bibinfo{person}{Konstantinos Rematas}, \bibinfo{person}{Ira
  Kemelmacher-Shlizerman}, \bibinfo{person}{Brian Curless}, {and}
  \bibinfo{person}{Steve Seitz}.} \bibinfo{year}{2018}\natexlab{}.
\newblock \showarticletitle{Soccer on Your Tabletop}. In
  \bibinfo{booktitle}{\emph{IEEE Conf. Comput. Vis. Pattern Recog.}}
\newblock


\bibitem[\protect\citeauthoryear{Richardt, Kim, Valgaerts, and
  Theobalt}{Richardt et~al\mbox{.}}{2016}]%
        {richardt2016dense}
\bibfield{author}{\bibinfo{person}{Christian Richardt},
  \bibinfo{person}{Hyeongwoo Kim}, \bibinfo{person}{Levi Valgaerts}, {and}
  \bibinfo{person}{Christian Theobalt}.} \bibinfo{year}{2016}\natexlab{}.
\newblock \showarticletitle{Dense wide-baseline scene flow from two handheld
  video cameras}. In \bibinfo{booktitle}{\emph{2016 Fourth International
  Conference on 3D Vision (3DV)}}. IEEE, \bibinfo{pages}{276--285}.
\newblock


\bibitem[\protect\citeauthoryear{Russell, Yu, and Agapito}{Russell
  et~al\mbox{.}}{2014}]%
        {russell2014video}
\bibfield{author}{\bibinfo{person}{Chris Russell}, \bibinfo{person}{Rui Yu},
  {and} \bibinfo{person}{Lourdes Agapito}.} \bibinfo{year}{2014}\natexlab{}.
\newblock \showarticletitle{Video pop-up: Monocular 3d reconstruction of
  dynamic scenes}. In \bibinfo{booktitle}{\emph{Eur. Conf. Comput. Vis.}}
  \bibinfo{pages}{583--598}.
\newblock


\bibitem[\protect\citeauthoryear{Sch\"{o}nberger and Frahm}{Sch\"{o}nberger and
  Frahm}{2016}]%
        {schoenberger2016sfm}
\bibfield{author}{\bibinfo{person}{Johannes~Lutz Sch\"{o}nberger} {and}
  \bibinfo{person}{Jan-Michael Frahm}.} \bibinfo{year}{2016}\natexlab{}.
\newblock \showarticletitle{Structure-from-Motion Revisited}. In
  \bibinfo{booktitle}{\emph{Conference on Computer Vision and Pattern
  Recognition (CVPR)}}.
\newblock


\bibitem[\protect\citeauthoryear{Sch\"{o}nberger, Zheng, Pollefeys, and
  Frahm}{Sch\"{o}nberger et~al\mbox{.}}{2016}]%
        {schoenberger2016mvs}
\bibfield{author}{\bibinfo{person}{Johannes~Lutz Sch\"{o}nberger},
  \bibinfo{person}{Enliang Zheng}, \bibinfo{person}{Marc Pollefeys}, {and}
  \bibinfo{person}{Jan-Michael Frahm}.} \bibinfo{year}{2016}\natexlab{}.
\newblock \showarticletitle{Pixelwise View Selection for Unstructured
  Multi-View Stereo}. In \bibinfo{booktitle}{\emph{European Conference on
  Computer Vision (ECCV)}}.
\newblock


\bibitem[\protect\citeauthoryear{Seitz, Curless, Diebel, Scharstein, and
  Szeliski}{Seitz et~al\mbox{.}}{2006}]%
        {seitz2006comparison}
\bibfield{author}{\bibinfo{person}{Steven~M Seitz}, \bibinfo{person}{Brian
  Curless}, \bibinfo{person}{James Diebel}, \bibinfo{person}{Daniel
  Scharstein}, {and} \bibinfo{person}{Richard Szeliski}.}
  \bibinfo{year}{2006}\natexlab{}.
\newblock \showarticletitle{A comparison and evaluation of multi-view stereo
  reconstruction algorithms}. In \bibinfo{booktitle}{\emph{2006 IEEE computer
  society conference on computer vision and pattern recognition (CVPR'06)}},
  Vol.~\bibinfo{volume}{1}. IEEE, \bibinfo{pages}{519--528}.
\newblock


\bibitem[\protect\citeauthoryear{Simon, Valmadre, Matthews, and Sheikh}{Simon
  et~al\mbox{.}}{2017}]%
        {Simon2017KroneckerMarkovPF}
\bibfield{author}{\bibinfo{person}{Tomas Simon}, \bibinfo{person}{Jack
  Valmadre}, \bibinfo{person}{Iain~A. Matthews}, {and} \bibinfo{person}{Yaser
  Sheikh}.} \bibinfo{year}{2017}\natexlab{}.
\newblock \showarticletitle{{Kronecker-Markov Prior for Dynamic 3D
  Reconstruction}}.
\newblock \bibinfo{journal}{\emph{IEEE Trans. Pattern Anal. Mach. Intell.}}
  \bibinfo{volume}{39} (\bibinfo{year}{2017}), \bibinfo{pages}{2201--2214}.
\newblock


\bibitem[\protect\citeauthoryear{Taniai, Sinha, and Sato}{Taniai
  et~al\mbox{.}}{2017}]%
        {Taniai2017}
\bibfield{author}{\bibinfo{person}{Tatsunori Taniai},
  \bibinfo{person}{Sudipta~N. Sinha}, {and} \bibinfo{person}{Yoichi Sato}.}
  \bibinfo{year}{2017}\natexlab{}.
\newblock \showarticletitle{{Fast Multi-frame Stereo Scene Flow with Motion
  Segmentation}}. In \bibinfo{booktitle}{\emph{IEEE Conference on Computer
  Vision and Pattern Recognition (CVPR)}}. \bibinfo{pages}{6891--6900}.
\newblock


\bibitem[\protect\citeauthoryear{Teed and Deng}{Teed and Deng}{2020}]%
        {teed2020raft}
\bibfield{author}{\bibinfo{person}{Zachary Teed} {and} \bibinfo{person}{Jia
  Deng}.} \bibinfo{year}{2020}\natexlab{}.
\newblock \showarticletitle{Raft: Recurrent all-pairs field transforms for
  optical flow}. In \bibinfo{booktitle}{\emph{European Conference on Computer
  Vision}}. Springer, \bibinfo{pages}{402--419}.
\newblock


\bibitem[\protect\citeauthoryear{Torresani, Hertzmann, and Bregler}{Torresani
  et~al\mbox{.}}{2008}]%
        {torresani2008nrsfm}
\bibfield{author}{\bibinfo{person}{Lorenzo Torresani}, \bibinfo{person}{Aaron
  Hertzmann}, {and} \bibinfo{person}{Christoph Bregler}.}
  \bibinfo{year}{2008}\natexlab{}.
\newblock \showarticletitle{Nonrigid Structure-from-Motion: Estimating Shape
  and Motion with Hierarchical Priors}.
\newblock \bibinfo{journal}{\emph{IEEE transactions on pattern analysis and
  machine intelligence}}  \bibinfo{volume}{30} (\bibinfo{date}{06}
  \bibinfo{year}{2008}), \bibinfo{pages}{878--92}.
\newblock
\urldef\tempurl%
\url{https://doi.org/10.1109/TPAMI.2007.70752}
\showDOI{\tempurl}


\bibitem[\protect\citeauthoryear{Vo, Narasimhan, and Sheikh}{Vo
  et~al\mbox{.}}{2016}]%
        {vo2016spatiotemporal}
\bibfield{author}{\bibinfo{person}{Minh Vo}, \bibinfo{person}{Srinivasa~G
  Narasimhan}, {and} \bibinfo{person}{Yaser Sheikh}.}
  \bibinfo{year}{2016}\natexlab{}.
\newblock \showarticletitle{Spatiotemporal bundle adjustment for dynamic 3d
  reconstruction}. In \bibinfo{booktitle}{\emph{Proceedings of the IEEE
  Conference on Computer Vision and Pattern Recognition}}.
  \bibinfo{pages}{1710--1718}.
\newblock


\bibitem[\protect\citeauthoryear{Wang, Lucey, Perazzi, and Wang}{Wang
  et~al\mbox{.}}{2019}]%
        {wang2019web}
\bibfield{author}{\bibinfo{person}{Chaoyang Wang}, \bibinfo{person}{Simon
  Lucey}, \bibinfo{person}{Federico Perazzi}, {and} \bibinfo{person}{Oliver
  Wang}.} \bibinfo{year}{2019}\natexlab{}.
\newblock \showarticletitle{Web stereo video supervision for depth prediction
  from dynamic scenes}. In \bibinfo{booktitle}{\emph{2019 International
  Conference on 3D Vision (3DV)}}. IEEE, \bibinfo{pages}{348--357}.
\newblock


\bibitem[\protect\citeauthoryear{Wedel, Brox, Vaudrey, Rabe, Franke, and
  Cremers}{Wedel et~al\mbox{.}}{2011}]%
        {wedel2011stereoscopic}
\bibfield{author}{\bibinfo{person}{Andreas Wedel}, \bibinfo{person}{Thomas
  Brox}, \bibinfo{person}{Tobi Vaudrey}, \bibinfo{person}{Clemens Rabe},
  \bibinfo{person}{Uwe Franke}, {and} \bibinfo{person}{Daniel Cremers}.}
  \bibinfo{year}{2011}\natexlab{}.
\newblock \showarticletitle{Stereoscopic scene flow computation for 3D motion
  understanding}.
\newblock \bibinfo{journal}{\emph{Int. J. Comput. Vis.}}
  (\bibinfo{year}{2011}).
\newblock


\bibitem[\protect\citeauthoryear{Xian, Shen, Cao, Lu, Xiao, Li, and Luo}{Xian
  et~al\mbox{.}}{2018}]%
        {xian2018monocular}
\bibfield{author}{\bibinfo{person}{Ke Xian}, \bibinfo{person}{Chunhua Shen},
  \bibinfo{person}{Zhiguo Cao}, \bibinfo{person}{Hao Lu}, \bibinfo{person}{Yang
  Xiao}, \bibinfo{person}{Ruibo Li}, {and} \bibinfo{person}{Zhenbo Luo}.}
  \bibinfo{year}{2018}\natexlab{}.
\newblock \showarticletitle{Monocular relative depth perception with web stereo
  data supervision}. In \bibinfo{booktitle}{\emph{Proceedings of the IEEE
  Conference on Computer Vision and Pattern Recognition}}.
  \bibinfo{pages}{311--320}.
\newblock


\bibitem[\protect\citeauthoryear{Yang, Wang, Wang, Xu, and Nevatia}{Yang
  et~al\mbox{.}}{2018}]%
        {yang2018every}
\bibfield{author}{\bibinfo{person}{Zhenheng Yang}, \bibinfo{person}{Peng Wang},
  \bibinfo{person}{Yang Wang}, \bibinfo{person}{Wei Xu}, {and}
  \bibinfo{person}{Ram Nevatia}.} \bibinfo{year}{2018}\natexlab{}.
\newblock \showarticletitle{Every pixel counts: Unsupervised geometry learning
  with holistic 3d motion understanding}. In
  \bibinfo{booktitle}{\emph{Proceedings of the European Conference on Computer
  Vision (ECCV) Workshops}}. \bibinfo{pages}{0--0}.
\newblock


\bibitem[\protect\citeauthoryear{Yin and Shi}{Yin and Shi}{2018}]%
        {yin2018geonet}
\bibfield{author}{\bibinfo{person}{Zhichao Yin} {and} \bibinfo{person}{Jianping
  Shi}.} \bibinfo{year}{2018}\natexlab{}.
\newblock \showarticletitle{Geonet: Unsupervised learning of dense depth,
  optical flow and camera pose}. In \bibinfo{booktitle}{\emph{Proceedings of
  the IEEE conference on computer vision and pattern recognition}}.
  \bibinfo{pages}{1983--1992}.
\newblock


\bibitem[\protect\citeauthoryear{Yoon, Kim, Gallo, Park, and Kautz}{Yoon
  et~al\mbox{.}}{2020}]%
        {yoon2020novel}
\bibfield{author}{\bibinfo{person}{Jae~Shin Yoon}, \bibinfo{person}{Kihwan
  Kim}, \bibinfo{person}{Orazio Gallo}, \bibinfo{person}{Hyun~Soo Park}, {and}
  \bibinfo{person}{Jan Kautz}.} \bibinfo{year}{2020}\natexlab{}.
\newblock \showarticletitle{Novel view synthesis of dynamic scenes with
  globally coherent depths from a monocular camera}. In
  \bibinfo{booktitle}{\emph{Proceedings of the IEEE/CVF Conference on Computer
  Vision and Pattern Recognition}}. \bibinfo{pages}{5336--5345}.
\newblock


\bibitem[\protect\citeauthoryear{Zhou, Brown, Snavely, and Lowe}{Zhou
  et~al\mbox{.}}{2017}]%
        {zhou2017unsupervised}
\bibfield{author}{\bibinfo{person}{Tinghui Zhou}, \bibinfo{person}{Matthew
  Brown}, \bibinfo{person}{Noah Snavely}, {and} \bibinfo{person}{David~G
  Lowe}.} \bibinfo{year}{2017}\natexlab{}.
\newblock \showarticletitle{Unsupervised learning of depth and ego-motion from
  video}. In \bibinfo{booktitle}{\emph{Proceedings of the IEEE conference on
  computer vision and pattern recognition}}. \bibinfo{pages}{1851--1858}.
\newblock


\end{thebibliography}
\appendix
\section{Appendix}
\subsection{Instability of Analytic Scene Flow}
\label{appendix:analytic_sf}

Scene flow computed directly from depth and optical flow is unstable when the angle between associated camera rays is small, as can be shown with the following derivation. Please see \cite{park20103d} for further explanation of the multi-view geometry. Under analytical sceneflow, we can rewrite $\prior_i(x)$ as:
\begin{equation}
    \begin{array}{ll}
    \prior_i(x) & = |\worldof{i+1}{x_1} - \worldof{i}{x} - \worldof{i+2}{x_2}+\worldof{i+1}{x_1}|_2\\
    & = |\worldof{i}{x} - 2\worldof{i+1}{x_1} + \worldof{i+2}{x_2}|_2,
    \end{array}
\end{equation}
where $x_1= x+\flow{i}{i+1}(x),\ x_2 = x_1 + \flow{i+1}{i+2}(x_1)$. We can rewrite $\worldof{i}{x}$ as $r_i(x)d^r_i(x)+t_i$, where $r_i(x)$ is the camera ray direction vector at $x$,  $d^r_i(x)$ the ray depth at $x$ and $t_i$ the camera translation vector. To simplify notation, we write $r(x)$ as $r_0$, $r_{i+1}(x_1)$ as $r_1$ and $r_{i+2}(x_2)$ as $r_2$. With similar rewriting for $d_i^r$ and $t_i$, we can rewrite $\prior_i(x)$ as :
\begin{equation}
    \begin{array}{l}
    \prior_i(x) = |r_0d^r_0+t_0 - 2r_1d^r_1-2t_1+r_2d^r_2+t_2|_2
    \end{array}
\end{equation}
Optimizing $\prior_i(x)$ then becomes solving $Rd=t$ under least squares, where $r_1$,$r_2$ and $r_3$ are columns of $R$, $d = [d^r_0, -2d^r_1, d^r_2]$ and $t=2t_1-t_2-t_0$. This problem is ill-posed when $R$ is poorly conditioned, which happens when the angles between $r_1$, $r_2$ and $r_3$ are small, for example when the object's and camera's motion align over the span of $3$ frames.

\end{document}